\RequirePackage{fix-cm}
\documentclass[11pt]{academic_template}
\usepackage[T1]{fontenc}
\usepackage{amsmath,amssymb,amsthm,booktabs,tabularx,array,graphicx,xcolor}
\usepackage{colortbl}
\usepackage{longtable}
\usepackage{microtype}
\usepackage{hyperref,url,placeins,caption}
\DeclareCaptionFont{compact}{\fontsize{8.5}{9.5}\selectfont}
\usepackage{fvextra}
\usepackage{enumitem}
\usepackage{tcolorbox}
\tcbuselibrary{skins,breakable}
\settoprulethickness{2.5pt}
\setbottomrulethickness{1.5pt}
\abstractboxon
\setabstractframecolor{gray}
\setabstractbgcolor{gray!10}
\renewcommand{\abstractinfont}{\fontsize{9}{10.8}\selectfont\linespread{1.0}\selectfont}
\setcitestyle{numbers,square}
\hypersetup{colorlinks=true,citecolor=themeblue,linkcolor=themeblue,urlcolor=themeblue,
  pdftitle={WorldFailBench: Do World Models Understand Policy Failures?},
  pdfauthor={Zheng Zhang, Xiaoyang Tang, Shuaiqi Cheng, Jiakai Huang, Zhongbo Wu,
    Ziyi Zhang, Yuan Si, Zeyu Liu, Weibo Gao, Shunyu Liu, Shan You, Fei Wang,
    Xiaogang Wang, Dacheng Tao}}
\setlogoheight{10mm}
\setlogotolineshift{0mm}
\fancypagestyle{firststyle}{%
  \fancyhf{}%
  \fancyhead[R]{%
    \raisebox{\dimexpr-\height+\logoverticalshift-1mm\relax}[0pt][0pt]{%
      \includegraphics[height=\logoheight]{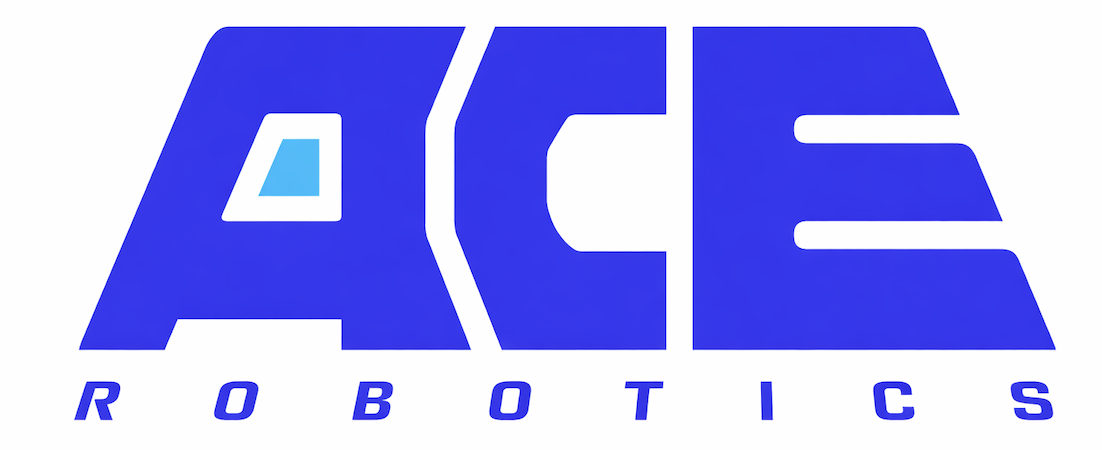}}}%
  \fancyfoot[L]{%
    \raisebox{18pt}[0pt][0pt]{%
      \fontsize{9}{11}\selectfont
      \textsuperscript{*} Equal Contribution.}}%
  \fancyfoot[C]{\thepage}}
\newcommand{\bench}{\textup{WorldFailBench}}
\newcommand{\NA}{\textemdash}

\graphicspath{{figures/}}
\title{\textup{WorldFailBench: Do World Models Understand Policy Failures?}}
\renewcommand{\authorfont}{\fontsize{11.5}{14.5}\selectfont}
\author[*]{Zheng Zhang}
\author[*]{Xiaoyang Tang}
\author{Shuaiqi Cheng}
\author{Jiakai Huang}
\author{Zhongbo Wu}
\author{Ziyi Zhang}
\author{Yuan Si}
\author{Zeyu Liu}
\author{Weibo Gao}
\author{Shunyu Liu}
\author{Shan You}
\author{Fei Wang}
\author{Xiaogang Wang}
\author{Dacheng Tao}
\abstract{
Robot policy evaluation informs deployment decisions and policy improvement, but physical testing and simulator preparation require substantial effort. World models offer a complementary route through closed-loop interaction in generated environments. Recent benchmarks have demonstrated their potential as policy evaluators through outcome-based evaluation (i.e., assessing task success or failure). However, such evaluation does not establish whether world models faithfully simulate how policies fail. Misrepresenting failures can lead to incorrect diagnoses of policy weaknesses and make policies appear ready for deployment even when small scene changes cause them to fail. We therefore ask: do world models faithfully simulate how and at which stages policies fail, and under what conditions they transition from success to failure? To answer this question, we introduce \bench{} with three evaluation tasks. Failure reason tests whether models reproduce the observable errors policies make. Failure stage assesses whether failures concentrate at the same stages of full-task execution. Failure boundary examines whether models reproduce the conditions under which policies transition from success to failure under controlled scene changes in simulation. The benchmark covers manipulation tasks in the LIBERO and RoboTwin simulation environments and real-world manipulation tasks performed by the G1 humanoid robot. Experiments across seven world models reveal substantial gaps in their understanding of policy failures, providing guidance for developing more reliable world models for policy evaluation.
}
\checkdata[Code]{\url{https://github.com/DAXIAORobotics/WorldFailBench}}
\checkdata[Hugging Face]{\url{https://huggingface.co/datasets/ACERobotics/WorldFailBench}}
\begin{document}
\raggedbottom
\setlength{\emergencystretch}{1em}
\maketitle
\clearpage
\begin{spacing}{0.9}
\tableofcontents
\end{spacing}
\clearpage

\section{Introduction}
Generalist robot policies increasingly support diverse manipulation tasks~\citep{team2024octo,pmlr-v270-kim25c,black2024pi_0}, with advances in multimodal prompting, bimanual control, and hierarchical reasoning~\citep{pmlr-v202-jiang23b,ICLR2025_49f80e4d,pmlr-v267-wen25g,ICLR2025_3bfee3bc}. Reliable deployment requires repeated evaluation across tasks and operating conditions. Physical testing occupies robot hardware and requires resets and supervision~\citep{pmlr-v305-atreya25a,pmlr-v305-zhou25a}, while traditional simulation requires scene construction and adaptation to the target environment~\citep{pmlr-v270-li25c}. Learned world models offer a complementary route: predicting future observations conditioned on robot actions~\citep{ICLR2024_c4d66eae,NEURIPS2024_7dbb5bfa} enables closed-loop policy evaluation before physical deployment~\citep{quevedo2026worldgym,guo2026ctrl}. To serve as reliable policy evaluators, they should faithfully simulate policy behavior in the target environments.

Recent benchmarks, including WorldArena and WMBench, systematically assess the reliability of world models as policy evaluators in simulated and real-world settings~\citep{shang2026worldarena,team2026gigaworld}. A common approach is \emph{outcome-based evaluation}~\citep{li2025worldeval,quevedo2026worldgym,guo2026ctrl,li2026dworldeval}. Specifically, the same policies are evaluated in a world model and a \emph{reference environment}, either a conventional simulator or a real-world robot setup. Policy success rates in the two environments are compared through cross-policy correlations. Reported agreement with real-world policy performance provides encouraging evidence for the use of world models as policy evaluators~\citep{wang2026interactive,jeon2026roboworld}.

\begin{figure}[t]
\centering
\includegraphics[width=0.85\linewidth]{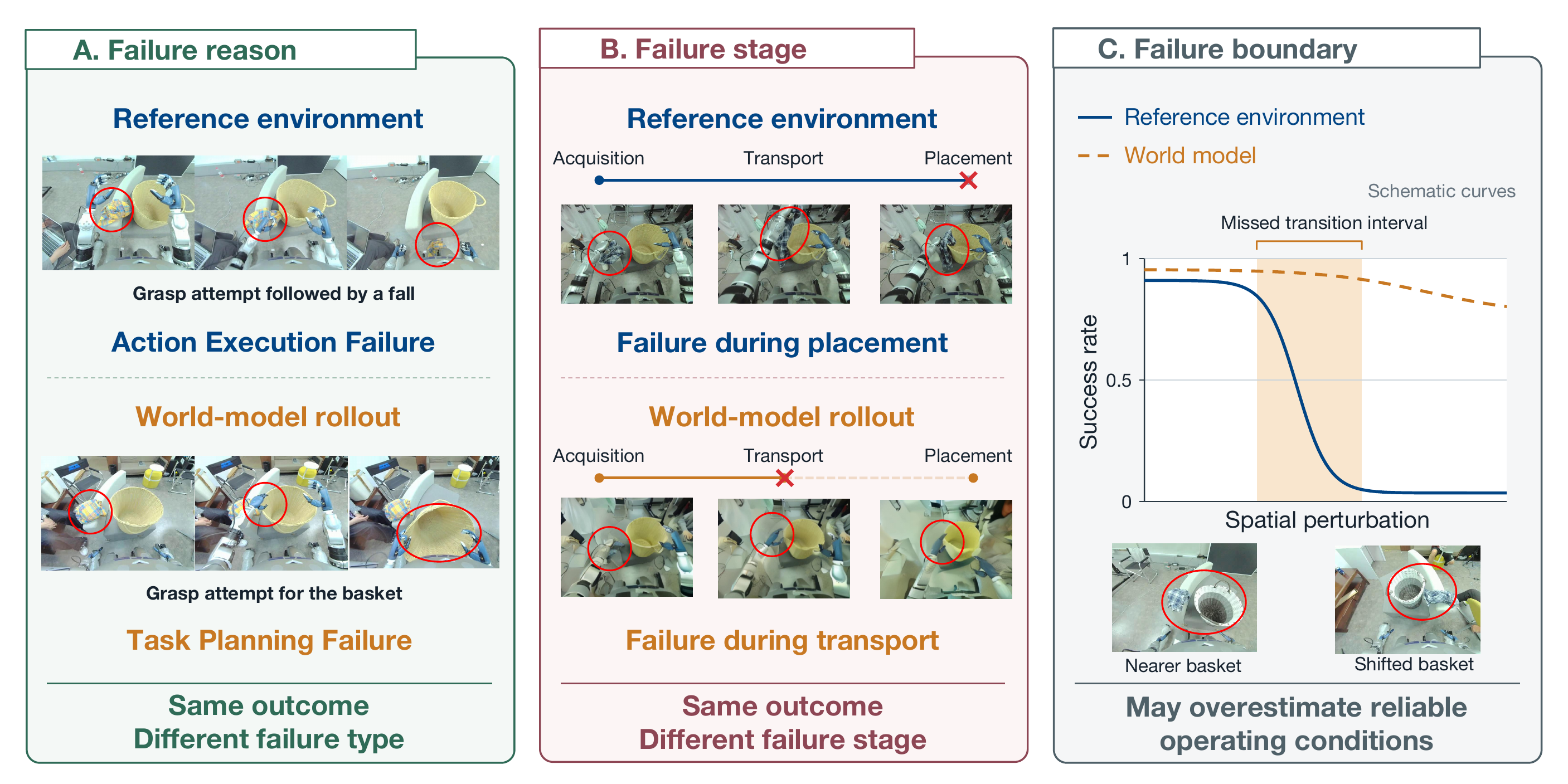}
\caption{Failure discrepancies in cloth collection. \textbf{(A)} Failure reason: dropping the cloth versus grasping the basket. \textbf{(B)} Failure stage: reference placement versus model transport. \textbf{(C)} Failure boundary: missed transitions under basket displacement.}
\label{fig:intro-errors}
\end{figure}

However, outcome-based evaluation does not establish whether world models faithfully simulate how a policy fails. Figure~\ref{fig:intro-errors} illustrates three discrepancies in cloth collection and their implications for policy assessment. \textbf{\mbox{(1) Failure reason.}} In panel A, reference execution attempts to grasp the cloth but drops it, indicating an action execution failure. The generated rollout instead attempts to grasp the basket, choosing the wrong target and indicating a \mbox{task planning failure}. Although both executions fail, this mismatch can misattribute an execution problem to task planning and direct policy improvement toward the wrong component. \textbf{\mbox{(2) Failure stage.}} In panel B, reference execution fails during placement, whereas the model predicts failure during transport. This misrepresents where execution breaks down despite agreement on the final outcome. \textbf{\mbox{(3) Failure boundary.}} Panel C illustrates how a model can continue to predict high success rates after basket displacement causes reference success to collapse. Missing this success-to-failure transition can overestimate the conditions under which a policy operates reliably. We therefore ask: \emph{do world models faithfully simulate how and at which stages policies fail, and under what conditions they transition from success to failure during closed-loop interaction?}

To answer this question, we introduce \bench{} with three complementary evaluation tasks. \textbf{Failure reason} asks whether a world model reproduces the kinds of observable errors a policy makes in the reference environment. We compare each policy's type-specific failure frequencies over all evaluation episodes and the correlation of these frequencies across policies in the two environments. \textbf{Failure stage} asks whether failures concentrate at the same stages of full-task execution as in the reference environment. We compare each policy's stage-specific failure frequencies over the same episodes and their cross-policy correlation between the two environments. \textbf{Failure boundary} asks whether the model captures the conditions under which a policy transitions from success to failure. In simulation, we select pairs of nearby conditions with a drop in reference success rate and test whether the model reproduces the direction and magnitude of that drop. The benchmark spans manipulation tasks in the LIBERO~\citep{liu2023libero} and RoboTwin~\citep{chen2025robotwin} simulation environments and real-world manipulation tasks performed by the G1 humanoid robot. Table~\ref{tab:benchmark-comparison} compares benchmark designs.

Experiments with seven world models reveal diagnostic gaps that outcome-based evaluation does not expose. On RoboTwin, average errors in predicted planning and execution failure frequencies range from 15.9 to 33.5 percentage points. All seven models show weak stage-specific policy correlations, including the model that best tracks policy success rates. Under scene changes, the models also fail to reliably capture success-to-failure transitions: 95.2\% of their responses on the tested RoboTwin boundary pairs understate the magnitude of reference success-rate changes. Our contributions are:

\begin{samepage}
\begin{itemize}
\setlength{\itemsep}{0pt}
\setlength{\parskip}{0pt}
\item We introduce \bench{}, which evaluates world-model fidelity from three complementary perspectives: failure types, failure stages, and failure boundaries.
\item We collect reference and model rollouts with failure annotations in LIBERO and RoboTwin simulation and on the G1 robot, and construct controlled boundary pairs in simulation.
\item Extensive experiments with seven world models show that outcome-based agreement does not ensure faithful failure diagnosis or boundary prediction, providing guidance for developing more reliable \mbox{world models} for policy evaluation.
\end{itemize}
\end{samepage}

\begin{table}[t]
\centering
\setlength{\tabcolsep}{3.0pt}
\scriptsize
\caption{Comparison of benchmark designs. $\checkmark$: dedicated evaluation; $\triangle$: partial or indirect; $\times$: not evaluated; \NA: unverified. $\dagger$: subtask outcome agreement.}
\label{tab:benchmark-comparison}
\resizebox{\linewidth}{!}{%
\begin{tabular}{lccccccccc}
\toprule
 & Protocol & \multicolumn{3}{c}{Failure understanding} & \multicolumn{2}{c}{Environment} & \multicolumn{3}{c}{Embodiment}\\
\cmidrule(lr){2-2}\cmidrule(lr){3-5}\cmidrule(lr){6-7}\cmidrule(lr){8-10}
Benchmark & WM loop & Reason & Stage-level & Boundary & Sim. & Real & Single & Dual & Humanoid\\
\midrule
WorldArena~\citep{shang2026worldarena} & $\checkmark$ & $\times$ & $\times$ & $\times$ & $\checkmark$ & $\times$ & $\times$ & $\checkmark$ & $\times$\\
WMBench~\citep{team2026gigaworld} & $\checkmark$ & $\times$ & $\checkmark^{\dagger}$ & $\times$ & $\checkmark$ & $\checkmark$ & \NA & $\checkmark$ & \NA\\
WorldArena 2.0~\citep{shang2026worldarena2} & $\checkmark$ & $\times$ & $\times$ & $\times$ & $\checkmark$ & $\checkmark$ & $\checkmark$ & $\checkmark$ & $\times$\\
EWMBench~\citep{yue2025ewmbench} & $\times$ & $\triangle$ & $\triangle$ & $\times$ & $\times$ & $\checkmark$ & \NA & $\checkmark$ & \NA\\
WorldSimBench~\textnormal{(robot)}~\citep{pmlr-v267-qin25f} & $\times$ & $\times$ & $\triangle$ & $\times$ & $\checkmark$ & $\times$ & $\checkmark$ & $\times$ & $\times$\\
RoboWM-Bench~\textnormal{(robot)}~\citep{jiang2026robowmbench} & $\times$ & $\checkmark$ & $\checkmark$ & $\times$ & $\checkmark$ & $\checkmark$ & $\checkmark$ & $\times$ & $\times$\\
WorldSimProbe~\citep{co2026worldsimprobe} & $\times$ & $\times$ & $\times$ & $\triangle$ & $\checkmark$ & $\times$ & $\checkmark$ & $\checkmark$ & $\times$\\
KineBench~\citep{liu2026kinebench} & $\times$ & $\times$ & $\times$ & $\times$ & $\checkmark$ & $\times$ & $\checkmark$ & $\times$ & $\times$\\
WorldEcho~\citep{chen2026worldecho} & $\times$ & $\triangle$ & $\times$ & $\times$ & $\checkmark$ & $\times$ & $\times$ & $\checkmark$ & $\times$\\
RoboEval~\citep{wang2025roboeval} & $\times$ & $\checkmark$ & $\checkmark$ & $\times$ & $\checkmark$ & $\times$ & $\times$ & $\checkmark$ & $\times$\\
MiraBench~\citep{yang2026mirabench} & $\times$ & $\checkmark$ & $\times$ & $\triangle$ & $\checkmark$ & $\checkmark$ & $\checkmark$ & $\checkmark$ & $\checkmark$\\
\midrule
\textbf{WorldFailBench} & $\checkmark$ & $\checkmark$ & $\checkmark$ & $\checkmark$ & $\checkmark$ & $\checkmark$ & $\checkmark$ & $\checkmark$ & $\checkmark$\\
\bottomrule
\end{tabular}%
}
\par\smallskip

\end{table}

\section{Related Work}
\paragraph{World-model-based policy evaluation and benchmarks.}
Learned world models increasingly serve as interactive surrogate environments for evaluating embodied policies. A recent survey distinguishes evaluation targets, protocols, metrics, and data when interpreting world-model benchmark results~\citep{zhang2026survey}. Systems such as WorldEval, WorldGym, and dWorldEval use action-conditioned video generation to estimate policy performance and compare policy rankings~\citep{li2025worldeval,quevedo2026worldgym,li2026dworldeval}, while GigaWorld-1 investigates how world-model architectures affect long-horizon, action-faithful evaluation~\citep{team2026gigaworld}. WorldArena provides unified benchmark protocols for embodied world models; WorldArena 2.0 extends evaluation to visuotactile inputs, policy optimization, and cross-embodiment sim-to-real settings~\citep{shang2026worldarena,shang2026worldarena2}.

Beyond policy-outcome evaluation, related benchmarks assess complementary aspects of world-model fidelity: WorldSimBench and EWMBench probe perceptual, motion, and semantic consistency~\citep{pmlr-v267-qin25f,yue2025ewmbench}; RoboWM-Bench tests whether generated interactions translate into executable behaviors~\citep{jiang2026robowmbench}; MiraBench analyzes reliability and over-optimism under failure-inducing actions~\citep{yang2026mirabench}; and RoboEval provides structured, stage-aware policy assessments~\citep{wang2025roboeval}. 
Table~\ref{tab:benchmark-comparison} summarizes these evaluation designs. Existing benchmarks provide outcome, executability, or recovery signals, but do not jointly assess agreement with reference execution in observable failure types, decisive failure stages, and local success-to-failure boundaries. WorldFailBench evaluates these three aspects through paired reference and model rollouts and controlled boundary interventions.

\paragraph{Fine-grained robotic failure analysis.}
Diagnosing failure modes is essential for reliable policy deployment and interpretable robot learning. Prior methods explain manipulation failures through multimodal reasoning~\citep{liu2023reflect,ICLR2025_70a06501}, monitor runtime consistency and execution progress~\citep{pmlr-v270-agia25a}, or provide structured, stage-aware policy assessment~\citep{wang2025roboeval}. RoboReward and FailBench evaluate automated success detectors and reward models~\citep{lee2026roboreward,navasardyan2026failbench}. These approaches assess policies or automated judges against reference executions. WorldFailBench uses paired reference and model rollouts to evaluate how faithfully generative world models reproduce observable failure types, stage distributions, and local success-to-failure boundaries.

\section{WorldFailBench}
\label{sec:benchmark}
\label{sec:formulation}
\begin{figure}[!t]
\centering
\includegraphics[width=\linewidth]{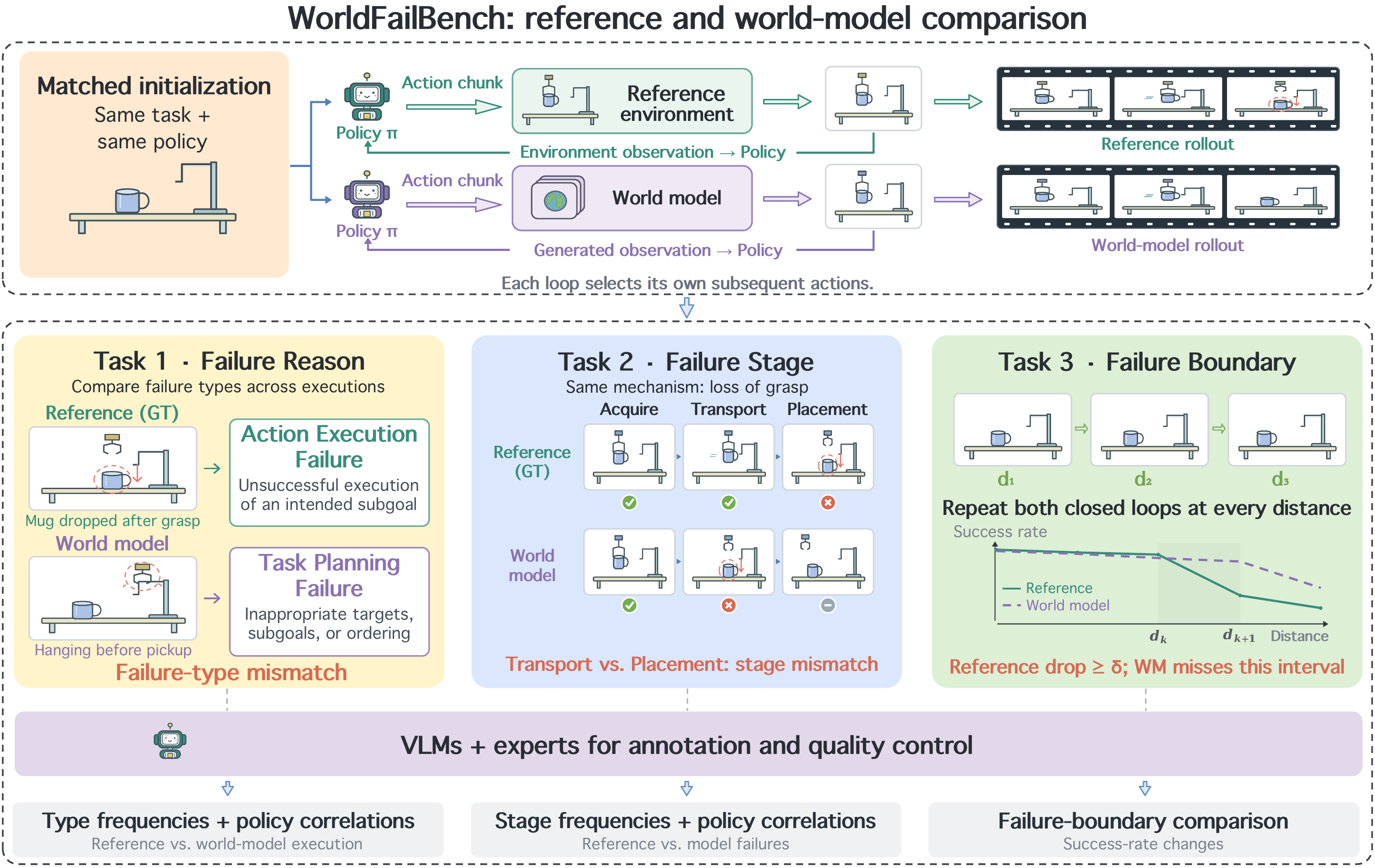}
\caption{Overview of \bench{}: comparing reference and world-model rollouts through failure reason, failure stage, and failure boundary.}
\label{fig:overview}
\end{figure}

\bench{} evaluates whether world models preserve policy failure behavior. Figure~\ref{fig:overview} summarizes the comparison: corresponding closed-loop executions provide failure-type and stage annotations, while controlled scene variations reveal success-to-failure conditions. We next describe the execution protocol, metrics, and benchmark construction.

\subsection{Closed-Loop Evaluation Framework}
\label{sec:closed_loop}
Closed-loop evaluation tests whether a policy exhibits consistent failure behavior when acting on generated rather than reference observations. At each step, the policy selects an action from its observation history, robot state, and task instruction; the world model predicts the next observation and returns it to the policy. Reference execution follows the same feedback process through a simulator or physical robot (Figure~\ref{fig:overview}). Each loop selects its own subsequent actions, allowing model-generated feedback to affect policy behavior.

To establish a common basis for comparing policy failures, we fix the task instruction, policy, and corresponding initial conditions across environments. Appendix~\ref{app:contract} provides further details of the rollout setup.

\subsection{Evaluation Tasks and Metrics}
\label{sec:protocol}
Failure Reason and Failure Stage compare failure frequencies and cross-policy correlations; Failure Boundary tests conditions under which policies transition from success to failure. Within each platform, all evaluated world models use the same tasks, policies, and retained reference episodes, with corresponding initial conditions. For task $t$ and policy $\pi$, $N_t(\pi)>0$ counts matched episodes with judged outcomes in each environment, including successes and failures. The environment index $r$ is $\mathrm{ref}$ for reference execution or $\mathrm{wm}$ for world-model execution. Vertical bars denote set size or absolute value, as appropriate.

\subsubsection{Task 1: Failure Reason}
\label{sec:task_reason}
Identifying the types of errors a policy makes helps determine which capabilities need improvement. Failure Reason evaluates whether world models reproduce type-specific failure frequencies and their variation across policies in reference execution.

Following REFLECT~\citep{liu2023reflect} and Guardian~\citep{pacaud2025guardian}, we distinguish two interpretable failure types: \emph{Task Planning Failure} (inappropriate targets, subgoals, or ordering) and \emph{Action Execution Failure} (unsuccessful execution of an intended subgoal). These types form the TFE set $\mathcal C$. Each failed trajectory receives one primary type; unclassifiable confirmed failures are labeled \emph{Indeterminate}. Their episodes remain in $N_t(\pi)$. Appendix~\ref{app:failure_categories} details annotation.

\paragraph{(1) Type Frequency Error (TFE).}
TFE compares failure-type frequencies for a given policy. For type $c\in\mathcal C$, let $n^{\mathrm{type}}_{r,t}(\pi,c)$ count failed episodes carrying that label. Its frequency and task-level error are:
\begin{equation}
 f^{\mathrm{type}}_{r,t}(\pi,c)=\frac{n^{\mathrm{type}}_{r,t}(\pi,c)}{N_t(\pi)},\qquad
 \mathrm{TFE}_t(\pi)=\frac{1}{|\mathcal C|}\sum_{c\in\mathcal C}
 \left|f^{\mathrm{type}}_{\mathrm{ref},t}(\pi,c)-f^{\mathrm{type}}_{\mathrm{wm},t}(\pi,c)\right|.
 \label{eq:type_frequency_error}
\end{equation}
Lower TFE indicates closer agreement. Unobserved types have zero frequency; task-level scores are averaged equally for each policy.

\goodbreak
\begin{samepage}
\paragraph{(2) Type-specific Policy Correlation (TPC).}
TPC tests whether a world model preserves cross-policy differences in failure-type frequencies. Following WorldArena~\citep{shang2026worldarena}, we use Pearson correlation across the same $J$ policies $\pi_1,\ldots,\pi_J$. For task $t$ and type $c$, $\mathbf f^{\mathrm{type}}_{r,t}(c)=(f^{\mathrm{type}}_{r,t}(\pi_j,c))_{j=1}^{J}$ lists one frequency per policy in environment $r$. Let $\mathcal C_t\subseteq\mathcal C$ contain the types with defined correlations in task $t$; $|\mathcal C_t|$ is their count. We compute each type's correlation and average them within the task:
\begin{equation}
 \mathrm{TPC}_t(c)=\operatorname{Pearson}
 \bigl(\mathbf f^{\mathrm{type}}_{\mathrm{ref},t}(c),\mathbf f^{\mathrm{type}}_{\mathrm{wm},t}(c)\bigr),\quad
 \mathrm{TPC}_t=\frac{1}{|\mathcal C_t|}
 \sum_{c\in\mathcal C_t}\mathrm{TPC}_t(c).
 \label{eq:category_ranking}
\end{equation}
\par
\end{samepage}
Pearson correlation is undefined when either input vector has zero variance. Overall TPC averages task scores with at least one defined type correlation equally. Higher TPC indicates stronger positive linear agreement across policies.

\subsubsection{Task 2: Failure Stage}
\label{sec:task_stage}
Failure Stage evaluates whether world models preserve where decisive failures occur during full-task execution and how their frequencies vary across policies. We divide task $t$ into subgoal-based stages $\mathcal S_t$ (Appendix~\ref{app:stage_decomposition}). We label the earliest decisive failure stage when it is identifiable.

\paragraph{(1) Stage Frequency Error (SFE).}
SFE measures the total absolute discrepancy in a policy's stage-failure frequencies. Let $n^{\mathrm{stage}}_{r,t}(\pi,s)$ count failed episodes labeled $s\in\mathcal S_t$. The frequency and task-level error are:
\begin{equation}
 f^{\mathrm{stage}}_{r,t}(\pi,s)=\frac{n^{\mathrm{stage}}_{r,t}(\pi,s)}{N_t(\pi)},\qquad
 \mathrm{SFE}_t(\pi)=\sum_{s\in\mathcal S_t}
 \left|f^{\mathrm{stage}}_{\mathrm{ref},t}(\pi,s)-f^{\mathrm{stage}}_{\mathrm{wm},t}(\pi,s)\right|.
 \label{eq:stage_frequency_error}
\end{equation}
Stage frequencies are computed over all evaluation episodes and reflect both stage reachability and failure after reaching a stage. Failures labeled \texttt{unknown} remain in the denominator but do not contribute to the named-stage counts. SFE lies in $[0,2]$, with lower values indicating closer agreement; task-level scores are averaged equally for each policy.

\paragraph{(2) Stage-specific Policy Correlation (SPC).}
SPC tests whether a world model preserves cross-policy differences in where decisive failures occur. For stage $s$, $\mathbf f^{\mathrm{stage}}_{r,t}(s)=(f^{\mathrm{stage}}_{r,t}(\pi_j,s))_{j=1}^{J}$ lists frequencies across the same policies. Let $\mathcal S_t^{\mathrm{corr}}\subseteq\mathcal S_t$ contain the stages with defined correlations in task $t$; $|\mathcal S_t^{\mathrm{corr}}|$ is their count. We compute each stage's Pearson correlation and average them within the task:
\begin{equation}
 \mathrm{SPC}_t(s)=\operatorname{Pearson}
 \bigl(\mathbf f^{\mathrm{stage}}_{\mathrm{ref},t}(s),\mathbf f^{\mathrm{stage}}_{\mathrm{wm},t}(s)\bigr),\quad
 \mathrm{SPC}_t=\frac{1}{|\mathcal S_t^{\mathrm{corr}}|}
 \sum_{s\in\mathcal S_t^{\mathrm{corr}}}\mathrm{SPC}_t(s).
 \label{eq:stage_macro}
\end{equation}
Overall SPC averages task scores with at least one defined stage correlation equally. Success-rate correlation applies the same procedure to success rates.

\subsubsection{Task 3: Failure Boundary}
\label{sec:task_boundary}
Identifying success-to-failure conditions can guide deployment and robustness improvements. Failure Boundary tests whether models reproduce these transitions for the same policy across nearby conditions differing in position, orientation, or size (Section~\ref{sec:construction}).

For a fixed task, policy, and base scene, $\mathcal B$ contains index pairs $(i,j)$ of selected scene conditions $x_i$ and $x_j$. Let $S_r(x)\in[0,1]$ be the empirical success rate at condition $x$ in environment $r$. The change is $\Delta_r(i,j)=S_r(x_i)-S_r(x_j)$, with endpoint order fixed during discovery and shared by both environments.

\paragraph{(1) Order Accuracy (OA).}
OA measures agreement in nonzero change directions:
\begin{equation}
 \mathrm{OA}=\frac{1}{|\mathcal B|}\sum_{(i,j)\in\mathcal B}
 \mathbf1[\Delta_{\mathrm{ref}}(i,j)\Delta_{\mathrm{wm}}(i,j)>0].
 \label{eq:oa}
\end{equation}
Here $\mathbf1[\cdot]$ equals one for a true condition and zero otherwise; ties score zero. Higher OA is better.

\paragraph{(2) Gap Error (GE).}
GE measures the error in success-rate changes:
\begin{equation}
 \mathrm{GE}=\frac{1}{|\mathcal B|}\sum_{(i,j)\in\mathcal B}
 |\Delta_{\mathrm{ref}}(i,j)-\Delta_{\mathrm{wm}}(i,j)|.
 \label{eq:ge}
\end{equation}
Lower GE is better. For each factor, we average over pairs within each task--policy--base-scene setting, then equally across settings.

\subsection{Benchmark Construction}
\label{sec:construction}
Benchmark construction comprises task and policy selection, rollout collection, boundary-pair selection, and annotation verification.

\begin{table}[!t]
\centering\small
\setlength{\tabcolsep}{3pt}
\caption{Benchmark composition and reference rollout allocation for failure-reason and failure-stage evaluation. Rollouts per task sum over that task's policies.}
\label{tab:coverage}
\begin{tabular*}{\linewidth}{@{\extracolsep{\fill}}llrrrrr@{}}
\toprule
Platform & \shortstack[l]{Reference\\environment} & Tasks & \shortstack{Policies /\\task} & \shortstack{Rollouts /\\task--policy} & \shortstack{Rollouts /\\task} & \shortstack{Total\\rollouts}\\
\midrule
LIBERO & Sim., single arm & 10 & 3 & 50 & 150 & 1,500\\
RoboTwin & Sim., bimanual & 20 & 5 & 50 & 250 & 5,000\\
G1 & Real, humanoid & 2 & 4 & 50 & 200 & 400\\
\bottomrule
\end{tabular*}
\end{table}

\begin{figure}[!t]
\centering
\includegraphics[width=\linewidth]{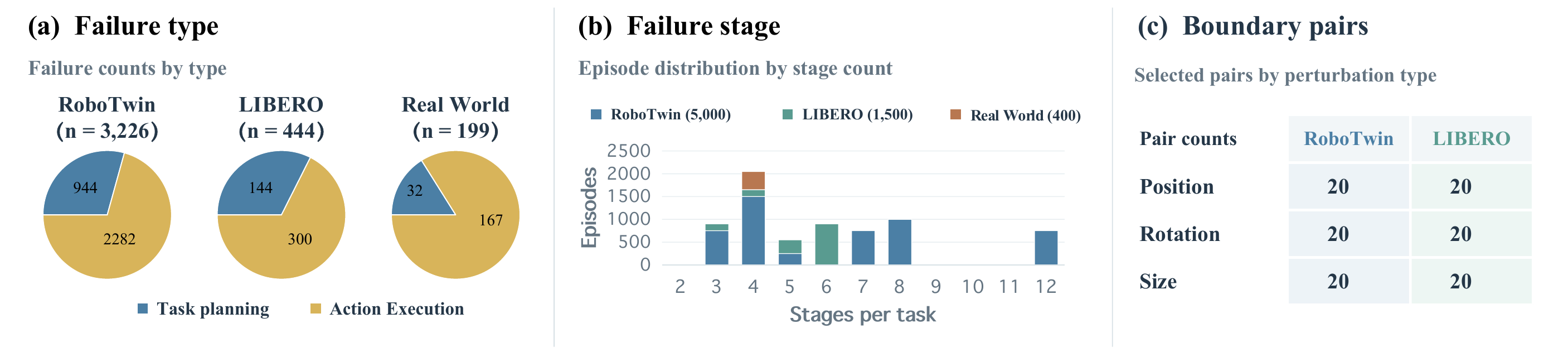}
\caption{Benchmark data distribution: (a) reference failure counts by type, with $n$ total reference failures per domain; (b) episode counts grouped by the number of named task stages; (c) boundary-pair counts by perturbation type. Real World denotes the real-world tasks on G1.}
\label{fig:data_distribution}
\end{figure}

\paragraph{Tasks and policies.}
Table~\ref{tab:coverage} summarizes single-arm LIBERO, bimanual RoboTwin, and real-world tasks on G1. LIBERO uses $\pi_{0.5}$, $\pi_{0.5}$-data-50, and $\pi_{0.5}$-data-20. The data-50 and data-20 variants use 50\% and 20\% of the policy fine-tuning data, respectively; $\pi_{0.5}$ uses the full dataset~\citep{intelligence2025pi05visionlanguageactionmodelopenworld}. RoboTwin additionally uses LingBot-VLA-v2 (LingBot)~\citep{wu2026foundation} and FastWAM~\citep{yuan2026fast}. Our real-world evaluation uses the G1 humanoid robot for two tasks: sweeping trash and collecting clothes. Each G1 task uses four task-specific policy versions~\citep{team2026kairos}, V1--V4. Policy checkpoints remain fixed across environments; Appendix~\ref{app:task_details} lists tasks and success criteria.

\paragraph{Failure-reason and failure-stage records.}
For each task--policy setting, we collect 50 reference rollouts from randomized scenes and corresponding world-model executions using Section~\ref{sec:closed_loop}. Both successful and failed rollouts enter annotation and frequency evaluation. Figure~\ref{fig:data_distribution}(a,b) shows reference failure counts by type and episode counts by the number of task stages, respectively.

\paragraph{Boundary pairs.}
On RoboTwin and LIBERO, we scan one scene factor---object position, orientation, or size---while holding other conditions fixed. Ten reference executions per condition identify adjacent pairs with a success-rate difference of at least $\delta=0.3$ (30 percentage points). We freeze 60 pairs per platform and their endpoint order, then independently re-estimate reference success using fresh random seeds (Appendix~\ref{app:boundary}).

For illustration, a LIBERO position scan in 2\,cm increments might show reference success falling from 90\% at a 4\,cm offset to 40\% at 6\,cm, qualifying the pair for selection. These hypothetical rates illustrate the criterion, rather than an additional measured pair. Each world model runs the same policy ten times per frozen endpoint. Pairs are not reselected or reordered after re-estimation. Figure~\ref{fig:data_distribution}(c) gives pair counts by factor.

\paragraph{Annotation and quality control.}
We use two annotation stages. \textbf{(1) Outcome annotation.} Doubao-Seed-2-1-Pro and ChatGPT 6 annotate video outcomes for reference and model rollouts, achieving 90.1\% agreement. One expert adjudicates all disagreements. We randomly audit agreed samples totaling 10\% of the full dataset, obtaining 97\% agreement with human judgments. For simulated reference executions, final outcomes follow native simulator predicates; video labels serve as verification. World-model rollouts and real-world G1 executions use adjudicated video labels. \textbf{(2) Failure annotation.} Two experts label the failure type and stage of every confirmed failure, achieving 92\% inter-annotator agreement before adjudication. A third annotator resolves all disagreements. Appendix~\ref{app:annotation} details annotation rules and outcome verification.

\section{Experiments}
\label{sec:experiments}
\subsection{Experimental Setup}
\label{sec:setup}
\paragraph{Evaluated world models.}
\looseness=-1 We evaluate Ctrl-World (Ctrl)~\citep{guo2026ctrl}, Cosmos-Predict-2.5 (Cosmos)~\citep{ali2025world}, BWM~\citep{bwm2026bwm}, DreamDojo~\citep{gao2026dreamdojo}, GE-Sim 2.0 (GE-Sim)~\citep{qiu2026ge}, A2World~\citep{huang2026learningtransferabledynamicspriors} and DW05~\citep{dexmal2026dw05} on RoboTwin. We evaluate the same models except BWM on LIBERO and G1; compatible BWM checkpoints and adaptation code were unavailable. Table~\ref{tab:coverage} summarizes the evaluation scope, and Appendix~\ref{app:configurations} gives configurations. The main results examine what failure diagnostics reveal alongside success-rate evaluation on RoboTwin; LIBERO and G1 results provide additional comparisons in Appendices~\ref{app:libero_type_results}--\ref{app:g1_failure_results}.

\looseness=-1 \textbf{Baseline.} Independent executions can differ even when the environment and policy are unchanged. To account for this variation, we rerun the same policies from the same initial scenes in RoboTwin and LIBERO, using the reference environment itself as an ideal simulator. Comparing the new and original executions with the same metrics gives the \textbf{Ref--Ref baseline}, which helps assess whether model discrepancies exceed the random variation observed in reference execution (Appendix~\ref{app:reference_repeatability}).

\FloatBarrier
\begin{table}[!t]
\centering\small
\setlength{\tabcolsep}{5pt}
\renewcommand{\arraystretch}{1.0}
\caption{Type Frequency Error (TFE) on RoboTwin ($\downarrow$). Policy columns average tasks; Mean weights the five policies equally. Bold marks the best world-model values.}
\label{tab:robotwin_tfe}
\begin{tabular}{@{}lrrrrrr@{}}
\toprule
Evaluator & $\pi_{0.5}$ & $\pi_{0.5}$-data-50 & $\pi_{0.5}$-data-20 & LingBot & FastWAM & Mean\\
\midrule
Ref--Ref & $0.059$ & $0.049$ & $0.068$ & $0.057$ & $0.072$ & $0.061$\\
\midrule
A2World & $0.212$ & $0.203$ & $0.217$ & $0.207$ & $0.216$ & $0.211$\\
BWM & $0.267$ & $0.230$ & $0.176$ & $0.262$ & $0.244$ & $0.236$\\
Cosmos & $0.422$ & $0.272$ & $0.295$ & $0.381$ & $0.304$ & $0.335$\\
Ctrl & $\mathbf{0.153}$ & $0.190$ & $0.141$ & $0.229$ & $\mathbf{0.142}$ & $0.171$\\
DW05 & $0.155$ & $\mathbf{0.166}$ & $\mathbf{0.136}$ & $\mathbf{0.164}$ & $0.174$ & $\mathbf{0.159}$\\
DreamDojo & $0.182$ & $0.214$ & $0.303$ & $0.189$ & $0.232$ & $0.224$\\
GE-Sim & $0.184$ & $0.174$ & $0.144$ & $0.189$ & $0.227$ & $0.184$\\
\bottomrule
\end{tabular}
\end{table}

\subsection{Failure-Type Understanding}
\label{sec:failure_type_results}
\begin{figure}[!t]
\centering
\includegraphics[trim=0 0 0 29bp,clip,width=0.85\linewidth]{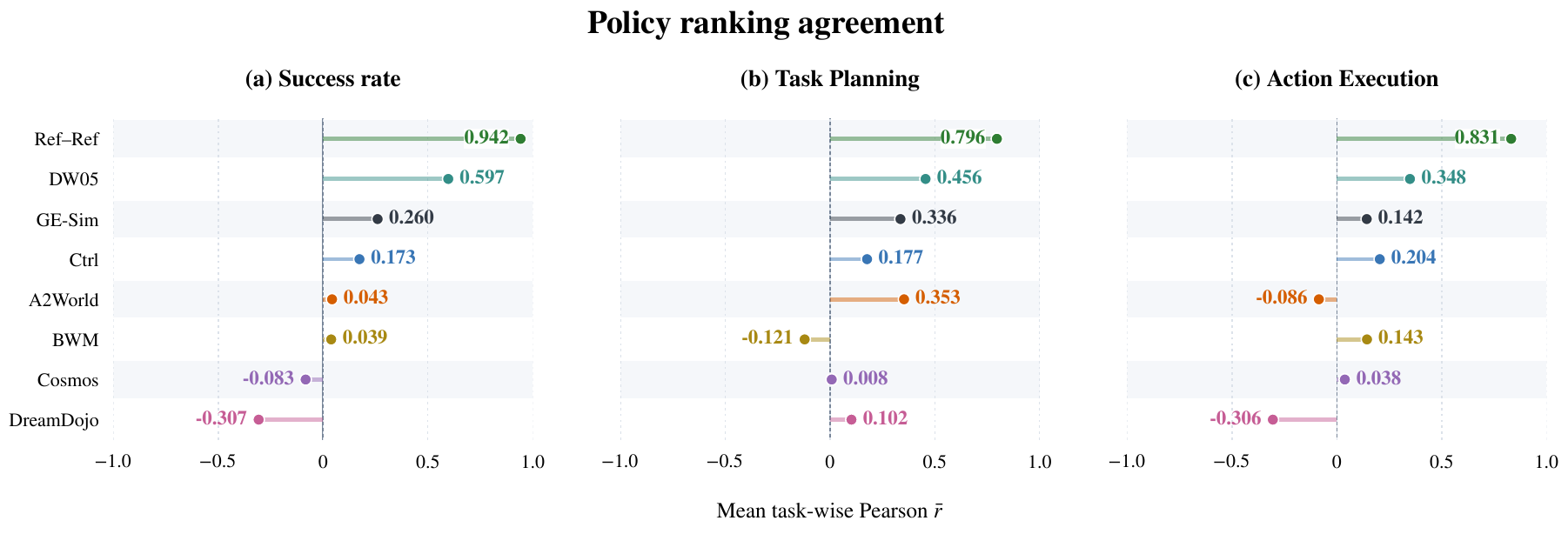}
\caption{RoboTwin mean within-task Pearson correlations for \textbf{(a)} success rate, \textbf{(b)} Task Planning, and \textbf{(c)} Action Execution. Ref--Ref is the reference-repeatability baseline; each correlation uses five policies.}
\label{fig:robotwin_type_policy_ranking}
\end{figure}

\looseness=-1 \textbf{RQ1: Does agreement in policy success rates extend to failure-type fidelity?} An evaluator must reproduce the kinds of failures behind the outcomes to support diagnosis of policy weaknesses. We compare success-rate and failure-type correlations across the same five RoboTwin policies (Figure~\ref{fig:robotwin_type_policy_ranking}), then measure errors in individual policies' failure frequencies using TFE (Table~\ref{tab:robotwin_tfe}).

\looseness=-1 \textbf{(1) Success-rate correlation does not establish consistent agreement across failure types.} DW05 has the highest success-rate correlation (0.597), but lower correlations for planning (0.456) and execution failures (0.348). Agreement also differs in direction across types: A2World correlates positively with reference planning failures (0.353) but negatively with execution failures ($-0.086$), while BWM shows the opposite pattern ($-0.121$ versus 0.143). The highest aggregate TPC is 0.402 (Table~\ref{tab:robotwin_type_rank}). Thus, an evaluator's ability to compare policies depends on which failure type is being assessed; its overall success-rate correlation does not describe this distinction.

\looseness=-1 \textbf{(2) World models misestimate how often policies exhibit planning and execution failures, distorting the diagnosis of their weaknesses.} Across the seven models, the average absolute error in these failure frequencies is 15.9--33.5 percentage points (Table~\ref{tab:robotwin_tfe}). Generated rollouts therefore give an inaccurate picture of how commonly each kind of failure occurs. For policy evaluation, these frequencies describe the prevalence of specific weaknesses, whereas the overall success rate records only how often tasks are completed.
\par

\begin{table}[!t]
\centering\small
\setlength{\tabcolsep}{5pt}
\renewcommand{\arraystretch}{1.0}
\caption{Stage Frequency Error (SFE) on RoboTwin ($\downarrow$). Policy columns average tasks; Mean averages policies within each task, then tasks. Frequencies use all valid paired episodes, including successes. Bold marks the best world-model values.}
\label{tab:robotwin_sfe}
\begin{tabular}{@{}lrrrrrr@{}}
\toprule
Evaluator & $\pi_{0.5}$ & $\pi_{0.5}$-data-50 & $\pi_{0.5}$-data-20 & LingBot & FastWAM & Mean\\
\midrule
Ref--Ref & $0.079$ & $0.104$ & $0.098$ & $0.097$ & $0.093$ & $0.094$\\
\midrule
A2World & $0.691$ & $0.880$ & $0.959$ & $0.613$ & $0.772$ & $0.783$\\
BWM & $0.682$ & $0.751$ & $0.808$ & $0.751$ & $0.766$ & $0.751$\\
Cosmos & $0.861$ & $0.760$ & $0.871$ & $0.853$ & $0.740$ & $0.817$\\
Ctrl & $0.525$ & $\mathbf{0.611}$ & $\mathbf{0.679}$ & $0.658$ & $0.691$ & $\mathbf{0.633}$\\
DW05 & $\mathbf{0.521}$ & $0.730$ & $0.818$ & $0.543$ & $\mathbf{0.663}$ & $0.655$\\
DreamDojo & $0.635$ & $0.626$ & $0.815$ & $0.545$ & $0.713$ & $0.667$\\
GE-Sim & $0.604$ & $0.823$ & $0.970$ & $\mathbf{0.517}$ & $0.692$ & $0.721$\\
\bottomrule
\end{tabular}
\end{table}

\subsection{Failure-Stage Understanding}
\label{sec:failure_stage_results}
\begin{figure}[!t]
\begin{minipage}[t]{0.435\linewidth}
\vspace{0pt}
\captionsetup{font=compact,position=top,skip=3pt}
\captionof{table}{Success-rate correlation and SPC on RoboTwin ($\uparrow$). Both average five-policy Pearson correlations over tasks; SPC averages stages within each task first. Bold marks the best world-model values.}
\label{tab:robotwin_policy_ranking}
\centering\fontsize{8.5}{10}\selectfont
\setlength{\tabcolsep}{3pt}
\renewcommand{\arraystretch}{1.12}
\begin{tabular*}{\linewidth}{@{\extracolsep{\fill}}lrr@{}}
\toprule
Evaluator & \shortstack{Success-rate\\correlation} & SPC\\
\midrule
Ref--Ref & $0.942$ & $0.903$\\
\midrule
A2World & $0.043$ & $-0.061$\\
BWM & $0.039$ & $0.045$\\
Cosmos & $-0.083$ & $-0.046$\\
Ctrl & $0.173$ & $\mathbf{0.095}$\\
DW05 & $\mathbf{0.597}$ & $0.052$\\
DreamDojo & $-0.307$ & $-0.123$\\
GE-Sim & $0.260$ & $-0.004$\\
\bottomrule
\end{tabular*}
\par\vspace{0pt}
\end{minipage}\hfill
\begin{minipage}[t]{0.545\linewidth}
\vspace{0pt}
\captionsetup{font=compact,position=top,skip=3pt}
\captionof{figure}{FastWAM on \texttt{blocks\_ranking\_size}. Cells show stage failure frequencies (\%) over all valid episodes for reference execution and seven world models.}
\label{fig:robotwin_stage_heatmap}
\centering
\includegraphics[trim=0 4.25bp 0 0,clip,width=\linewidth]{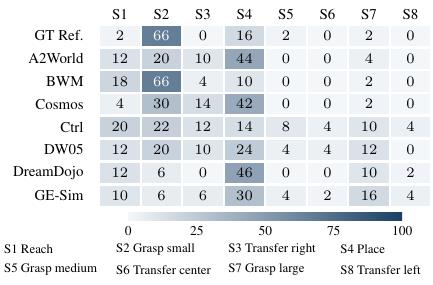}
\par\vspace{0pt}
\end{minipage}
\end{figure}

\looseness=-1 \textbf{RQ2: Do world models preserve where policies fail during full-task execution?} We test whether outcome agreement also supports comparisons of failure locations. Table~\ref{tab:robotwin_policy_ranking} contrasts success-rate correlation with SPC; Table~\ref{tab:robotwin_sfe} measures stage-frequency errors. We inspect the stage profiles in Figure~\ref{fig:robotwin_stage_heatmap} to explain the discrepancies.

\looseness=-1 \textbf{(1) The strongest outcome correlations still accompany weak agreement on where different policies fail.} DW05 has the highest success-rate correlation (0.597), yet its SPC is only 0.052; GE-Sim similarly drops from 0.260 to $-0.004$. Across all seven models, SPC ranges from $-0.123$ to 0.095. Ref--Ref attains 0.942 for success rate and 0.903 for stages, showing that both comparisons are substantially more reproducible between reference runs. The models' relative strength in outcome evaluation therefore does not carry over to comparing where policies fail during full-task execution.
\par

\looseness=-1 \textbf{(2) Generated rollouts misrepresent stage frequencies and can relocate the dominant failure.} Mean SFE ranges from 0.633 to 0.817. The FastWAM case in Figure~\ref{fig:robotwin_stage_heatmap} shows what this distortion means: in \texttt{blocks\_ranking\_size}, reference failures concentrate at Grasp small (66\%), with 16\% at Place, whereas DreamDojo reports 6\% and 46\%, respectively. This shift from grasping to placement appears across all five policies (Figure~\ref{fig:robotwin_stage_heatmap_all_policies}). The resulting diagnosis attributes the main breakdown in full-task execution to a different stage, information that an overall success rate does not provide.

\subsection{Failure-Boundary Understanding}
\label{sec:boundary_results}

\begin{figure}[!t]
\begin{minipage}[t]{0.53\linewidth}
\vspace{0pt}
\captionsetup{font=compact,position=top,skip=3pt}
\captionof{table}{RoboTwin failure-boundary results (OA $\uparrow$, GE $\downarrow$). World-model scores average settings within each factor. Ref--Ref reports reference repeatability; bold marks the best world model.}
\label{tab:boundary-response-fidelity}
\end{minipage}\hfill
\begin{minipage}[t]{0.45\linewidth}
\vspace{0pt}
\captionsetup{font=compact,position=top,skip=3pt}
\captionof{figure}{LingBot-VLA-v2 on \texttt{adjust\_bottle} (pair 032). The reference spans five yaw angles; models connect the highlighted $-60^\circ$/$-80^\circ$ endpoints.}
\label{fig:boundary-response-case}
\end{minipage}
\par\nointerlineskip
\begingroup
\sbox0{\begin{minipage}{0.53\linewidth}
\centering\fontsize{8.5}{10}\selectfont
\setlength{\tabcolsep}{2pt}
\begin{tabular*}{\linewidth}{@{\extracolsep{\fill}}lrrrrrr@{}}
\toprule
Evaluator & \multicolumn{2}{c}{Position} & \multicolumn{2}{c}{Orientation} & \multicolumn{2}{c}{Size} \\
\cmidrule(lr){2-3}\cmidrule(lr){4-5}\cmidrule(lr){6-7}
 & OA $\uparrow$ & GE $\downarrow$ & OA $\uparrow$ & GE $\downarrow$ & OA $\uparrow$ & GE $\downarrow$ \\
\midrule
Ref--Ref & 1.000 & 0.005 & 0.950 & 0.010 & 1.000 & 0.020 \\
\midrule
A2World & \textbf{0.685} & 0.515 & 0.410 & 0.619 & 0.455 & 0.576 \\
BWM & \textbf{0.685} & 0.529 & 0.256 & 0.614 & 0.364 & 0.589 \\
Cosmos & 0.583 & 0.587 & 0.321 & 0.617 & 0.394 & 0.648 \\
Ctrl & 0.611 & \textbf{0.459} & 0.359 & 0.719 & \textbf{0.758} & \textbf{0.500} \\
DreamDojo & 0.602 & 0.509 & 0.321 & 0.677 & 0.364 & 0.633 \\
DW05 & 0.556 & 0.503 & 0.308 & 0.658 & 0.288 & 0.636 \\
GE-Sim & 0.491 & 0.481 & \textbf{0.590} & \textbf{0.492} & 0.424 & 0.591 \\
\bottomrule
\end{tabular*}

\end{minipage}}
\usebox0\hfill
\begin{minipage}[c]{0.45\linewidth}
\centering
\includegraphics[width=\linewidth,height=\dimexpr\ht0+\dp0\relax,keepaspectratio]{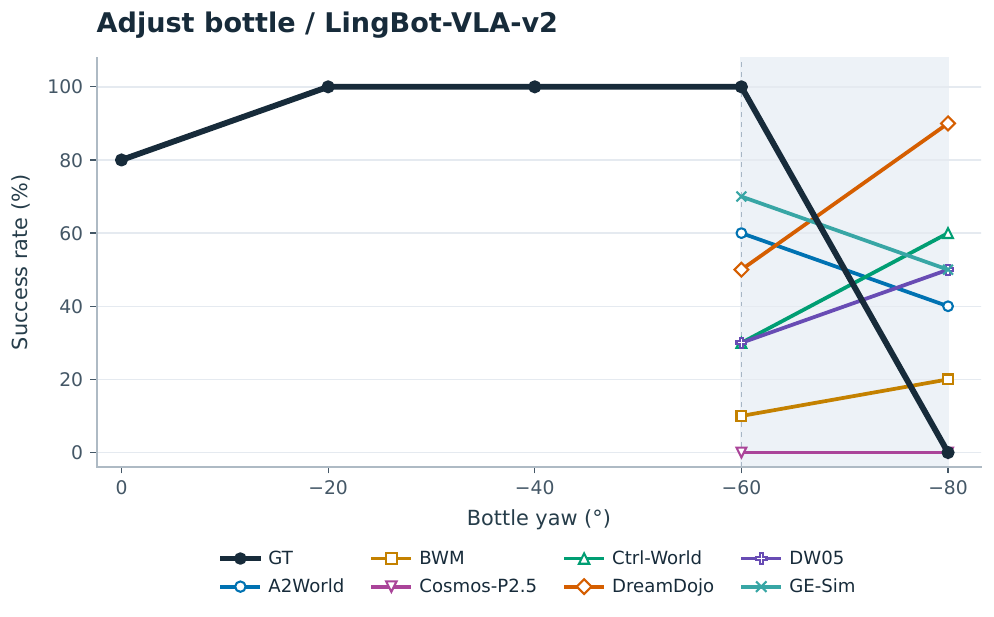}
\end{minipage}
\endgroup
\end{figure}

\looseness=-1 \textbf{RQ3: Do world models reproduce policy failure boundaries under scene changes?} We test whether generated rollouts identify which conditions reduce policy reliability and by how much. Table~\ref{tab:boundary-response-fidelity} reports OA and GE for seven world models and the Ref--Ref baseline on 60 RoboTwin boundary pairs spanning position, orientation, and size changes. Each pair holds the task, policy, and base scene fixed; Figure~\ref{fig:boundary-response-case} illustrates the resulting errors.

\FloatBarrier

\looseness=-1 \textbf{(1) World models misrepresent which scene changes degrade policy reliability and by how much.} Even the highest world-model OA for position, orientation, and size is only 68.5\%, 59.0\%, and 75.8\%, respectively (Table~\ref{tab:boundary-response-fidelity}). Thus, selecting the best model for each factor still leaves substantial disagreement about which nearby condition is less reliable. By comparison, Ref--Ref yields OA of 0.95--1.00 and GE of 0.005--0.020 across the three factors. Predicted changes also tend to be too small: across seven models on the same 60 RoboTwin boundary pairs, 400 of 420 model--pair comparisons (95.2\%) underestimate the absolute reference success-rate difference (Figure~\ref{fig:boundary-response-diagnostics}a--c). This compression makes policies appear less sensitive to scene changes than they are. Figure~\ref{fig:boundary-response-case} illustrates both errors: on \texttt{adjust\_bottle}, LingBot-VLA-v2's reference success falls from 100\% to 0\% as bottle yaw changes from $-60^\circ$ to $-80^\circ$. Four models predict an increase, and one predicts no change. Only A2World and GE-Sim identify the decline, but both predict a decrease of just 20 percentage points. Even a correct direction therefore leaves most of the reliability loss unrepresented in this case. For a policy evaluator, these errors can lead to overly optimistic robustness assessments by obscuring both the conditions that trigger failure and the severity of the resulting reliability loss.

\looseness=-1 \textbf{(2) Different scene changes expose distinct weaknesses in world-model policy evaluation.} Five of the seven models have their lowest OA on orientation changes, while DW05 and GE-Sim have their lowest OA on size changes (Table~\ref{tab:boundary-response-fidelity}). Ctrl-World illustrates this dependence: its OA falls from 75.8\% for size to 35.9\% for orientation, where it also has the highest GE among the seven models (0.719). GE-Sim exhibits the opposite pattern, with OA reaching 59.0\% for orientation but only 42.4\% for size. Testing size alone would miss Ctrl-World's difficulty with rotation, whereas testing orientation alone would miss GE-Sim's weaker size predictions. An evaluator that captures a policy's sensitivity to one scene factor may therefore misjudge its vulnerability to another. Failure-boundary evaluation makes this limitation explicit: it assesses whether a world model can support judgments about policy robustness under specific scene changes and identifies where its predictions require reference-environment verification before guiding further testing.

\section{Discussion and Conclusion}
We introduced WorldFailBench to evaluate whether world models preserve policy failure behavior beyond success rates. By comparing generated rollouts with reference executions, it measures agreement in failure types, failure stages, and responses to scene changes near failure boundaries.
Across the evaluated models, agreement in policy success rates did not consistently extend to failure diagnosis. Models with stronger outcome correlations could still misrepresent which failures occurred or when they occurred, while generated rollouts often understated the effects of scene changes on success. These findings show why failure fidelity matters when using world models to interpret policy behavior.

Our evaluation covers a finite set of policies, tasks, and boundary conditions, which limits the generality of these findings. The results characterize complete world-model systems, including their adapters and rollout schedules, and do not isolate individual architectural components. Failure labels describe observable behavior without establishing internal policy causes. Future work will extend the evaluation to broader settings and examine whether assessing failure fidelity improves policy training.

\FloatBarrier
\clearpage
\phantomsection
\addcontentsline{toc}{section}{References}
\bibliographystyle{unsrt}
\bibliography{references}

\begin{thebibliography}{10}

\bibitem{team2024octo}
Octo~Model Team, Dibya Ghosh, Homer Walke, Karl Pertsch, Kevin Black, Oier
  Mees, Sudeep Dasari, Joey Hejna, Tobias Kreiman, Charles Xu, et~al.
\newblock {Octo: An Open-Source Generalist Robot Policy}.
\newblock {\em arXiv preprint arXiv:2405.12213}, 2024.

\bibitem{pmlr-v270-kim25c}
Moo~Jin Kim, Karl Pertsch, Siddharth Karamcheti, Ted Xiao, Ashwin Balakrishna,
  Suraj Nair, Rafael Rafailov, Ethan~P Foster, Pannag~R Sanketi, Quan Vuong,
  Thomas Kollar, Benjamin Burchfiel, Russ Tedrake, Dorsa Sadigh, Sergey Levine,
  Percy Liang, and Chelsea Finn.
\newblock {OpenVLA: An Open-Source Vision-Language-Action Model}.
\newblock In Pulkit Agrawal, Oliver Kroemer, and Wolfram Burgard, editors, {\em
  Proceedings of The 8th Conference on Robot Learning}, volume 270 of {\em
  Proceedings of Machine Learning Research}, pages 2679--2713. PMLR, 06--09 Nov
  2025.

\bibitem{black2024pi_0}
Kevin Black, Noah Brown, Danny Driess, Adnan Esmail, Michael Equi, Chelsea
  Finn, Niccolo Fusai, Lachy Groom, Karol Hausman, Brian Ichter, et~al.
\newblock {$\pi_0$: A Vision-Language-Action Flow Model for General Robot
  Control}.
\newblock {\em arXiv preprint arXiv:2410.24164}, 2024.

\bibitem{pmlr-v202-jiang23b}
Yunfan Jiang, Agrim Gupta, Zichen Zhang, Guanzhi Wang, Yongqiang Dou, Yanjun
  Chen, Li~Fei-Fei, Anima Anandkumar, Yuke Zhu, and Linxi Fan.
\newblock {{VIMA}: Robot Manipulation with Multimodal Prompts}.
\newblock In Andreas Krause, Emma Brunskill, Kyunghyun Cho, Barbara Engelhardt,
  Sivan Sabato, and Jonathan Scarlett, editors, {\em Proceedings of the 40th
  International Conference on Machine Learning}, volume 202 of {\em Proceedings
  of Machine Learning Research}, pages 14975--15022. PMLR, 23--29 Jul 2023.

\bibitem{ICLR2025_49f80e4d}
Songming Liu, Lingxuan Wu, Bangguo Li, Hengkai Tan, Huayu Chen, Zhengyi Wang,
  Ke~Xu, Hang Su, and Jun Zhu.
\newblock {RDT-1B: a Diffusion Foundation Model for Bimanual Manipulation}.
\newblock In Y.~Yue, A.~Garg, N.~Peng, F.~Sha, and R.~Yu, editors, {\em
  International Conference on Learning Representations}, volume 2025, pages
  29982--30009, 2025.

\bibitem{pmlr-v267-wen25g}
Junjie Wen, Yichen Zhu, Minjie Zhu, Zhibin Tang, Jinming Li, Zhongyi Zhou,
  Xiaoyu Liu, Chaomin Shen, Yaxin Peng, and Feifei Feng.
\newblock {{D}iffusion{VLA}: Scaling Robot Foundation Models via Unified
  Diffusion and Autoregression}.
\newblock In Aarti Singh, Maryam Fazel, Daniel Hsu, Simon Lacoste-Julien, Felix
  Berkenkamp, Tegan Maharaj, Kiri Wagstaff, and Jerry Zhu, editors, {\em
  Proceedings of the 42nd International Conference on Machine Learning}, volume
  267 of {\em Proceedings of Machine Learning Research}, pages 66558--66574.
  PMLR, 13--19 Jul 2025.

\bibitem{ICLR2025_3bfee3bc}
Yi~Li, Yuquan Deng, Jesse Zhang, Joel Jang, Marius Memmel, Caelan Garrett,
  Fabio Ramos, Dieter Fox, Anqi Li, Abhishek Gupta, and Ankit Goyal.
\newblock {HAMSTER: Hierarchical Action Models for Open-World Robot
  Manipulation}.
\newblock In Y.~Yue, A.~Garg, N.~Peng, F.~Sha, and R.~Yu, editors, {\em
  International Conference on Learning Representations}, volume 2025, pages
  24040--24068, 2025.

\bibitem{pmlr-v305-atreya25a}
Pranav Atreya, Karl Pertsch, Tony Lee, Moo~Jin Kim, Arhan Jain, Artur
  Kuramshin, Cyrus Neary, Edward~S. Hu, Kanav Arora, Kirsty Ellis, Luca
  Macesanu, Matthew Leonard, Meedeum Cho, Ozgur Aslan, Shivin Dass, Tony Wang,
  Xingfang Yuan, Abhishek Gupta, Dinesh Jayaraman, Glen Berseth, Kostas
  Daniilidis, Roberto Mart\'{i}n-Mart\'{i}n, Youngwoon Lee, Percy Liang,
  Chelsea Finn, and Sergey Levine.
\newblock {RoboArena: Distributed Real-World Evaluation of Generalist Robot
  Policies}.
\newblock In Joseph Lim, Shuran Song, and Hae-Won Park, editors, {\em
  Proceedings of The 9th Conference on Robot Learning}, volume 305 of {\em
  Proceedings of Machine Learning Research}, pages 336--364. PMLR, 27--30 Sep
  2025.

\bibitem{pmlr-v305-zhou25a}
Zhiyuan Zhou, Pranav Atreya, You~Liang Tan, Karl Pertsch, and Sergey Levine.
\newblock {AutoEval: Autonomous Evaluation of Generalist Robot Manipulation
  Policies in the Real World}.
\newblock In Joseph Lim, Shuran Song, and Hae-Won Park, editors, {\em
  Proceedings of The 9th Conference on Robot Learning}, volume 305 of {\em
  Proceedings of Machine Learning Research}, pages 1997--2017. PMLR, 27--30 Sep
  2025.

\bibitem{pmlr-v270-li25c}
Xuanlin Li, Kyle Hsu, Jiayuan Gu, Oier Mees, Karl Pertsch, Homer~Rich Walke,
  Chuyuan Fu, Ishikaa Lunawat, Isabel Sieh, Sean Kirmani, Sergey Levine, Jiajun
  Wu, Chelsea Finn, Hao Su, Quan Vuong, and Ted Xiao.
\newblock {Evaluating Real-World Robot Manipulation Policies in Simulation}.
\newblock In Pulkit Agrawal, Oliver Kroemer, and Wolfram Burgard, editors, {\em
  Proceedings of The 8th Conference on Robot Learning}, volume 270 of {\em
  Proceedings of Machine Learning Research}, pages 3705--3728. PMLR, 06--09 Nov
  2025.

\bibitem{ICLR2024_c4d66eae}
Sherry Yang, Yilun Du, Seyed Ghasemipour, Jonathan Tompson, Leslie Kaelbling,
  Dale Schuurmans, and Pieter Abbeel.
\newblock {Learning Interactive Real-World Simulators}.
\newblock In B.~Kim, Y.~Yue, S.~Chaudhuri, K.~Fragkiadaki, M.~Khan, and Y.~Sun,
  editors, {\em International Conference on Learning Representations}, volume
  2024, pages 45210--45234, 2024.

\bibitem{NEURIPS2024_7dbb5bfa}
Jialong Wu, Shaofeng Yin, Ningya Feng, Xu~He, Dong Li, Jianye Hao, and
  Mingsheng Long.
\newblock {iVideoGPT: Interactive VideoGPTs are Scalable World Models}.
\newblock In A.~Globerson, L.~Mackey, D.~Belgrave, A.~Fan, U.~Paquet,
  J.~Tomczak, and C.~Zhang, editors, {\em Advances in Neural Information
  Processing Systems}, volume~37, pages 68082--68119. Curran Associates, Inc.,
  2024.

\bibitem{quevedo2026worldgym}
Julian Quevedo, Ansh~Kumar Sharma, Yixiang Sun, Varad Suryavanshi, Percy Liang,
  and Sherry Yang.
\newblock {{WorldGym}: World model as an environment for policy evaluation}.
\newblock In {\em International Conference on Learning Representations}, volume
  2026, pages 78932--78957, 2026.

\bibitem{guo2026ctrl}
Yanjiang Guo, Lucy Shi, Jianyu Chen, and Chelsea Finn.
\newblock {{Ctrl-World}: A controllable generative world model for robot
  manipulation}.
\newblock In {\em International Conference on Learning Representations}, volume
  2026, pages 6121--6138, 2026.

\bibitem{shang2026worldarena}
Yu~Shang, Zhuohang Li, Yiding Ma, Weikang Su, Xin Jin, Ziyou Wang, Lei Jin, Xin
  Zhang, Yinzhou Tang, Haisheng Su, et~al.
\newblock {{WorldArena}: A unified benchmark for evaluating perception and
  functional utility of embodied world models}.
\newblock {\em AI Open}, 7:208--226, 2026.

\bibitem{team2026gigaworld}
GigaWorld Team, Angyuan Ma, Boyuan Wang, Bohan Li, Chaojun Ni, Guo Li, Guan
  Huang, Guosheng Zhao, Hao Li, Hengtao Li, et~al.
\newblock {GigaWorld-1: A Roadmap to Build World Models for Robot Policy
  Evaluation}.
\newblock {\em arXiv preprint arXiv:2607.02642}, 2026.

\bibitem{li2025worldeval}
Yaxuan Li, Yichen Zhu, Junjie Wen, Chaomin Shen, and Yi~Xu.
\newblock {WorldEval: World Model as Real-World Robot Policies Evaluator}.
\newblock {\em arXiv preprint arXiv:2505.19017}, 2025.

\bibitem{li2026dworldeval}
Yaxuan Li, Zhongyi Zhou, Yefei Chen, Yaokai Xue, and Yichen Zhu.
\newblock {dWorldEval: Scalable Robotic Policy Evaluation via Discrete
  Diffusion World Model}.
\newblock {\em arXiv preprint arXiv:2604.22152}, 2026.

\bibitem{wang2026interactive}
Yixuan Wang, Rhythm Syed, Fangyu Wu, Mengchao Zhang, Aykut Onol, Jose
  Barreiros, Hooshang Nayyeri, Tony Dear, Huan Zhang, and Yunzhu Li.
\newblock {Interactive World Simulator for Robot Policy Training and
  Evaluation}.
\newblock {\em arXiv preprint arXiv:2603.08546}, 2026.

\bibitem{jeon2026roboworld}
Byeongguk Jeon, Seonghyeon Ye, JaeHyeok Doo, Sungdong Kim, Minjoon Seo,
  Hyungmok Son, and Kimin Lee.
\newblock {RoboWorld: Fast and Reliable Neural Simulators for Generalist Robot
  Policy Evaluation}.
\newblock {\em arXiv preprint arXiv:2607.01060}, 2026.

\bibitem{liu2023libero}
Bo~Liu, Yifeng Zhu, Chongkai Gao, Yihao Feng, Qiang Liu, Yuke Zhu, and Peter
  Stone.
\newblock {{LIBERO}: Benchmarking knowledge transfer for lifelong robot
  learning}.
\newblock {\em Advances in Neural Information Processing Systems},
  36:44776--44791, 2023.

\bibitem{chen2025robotwin}
Tianxing Chen, Zanxin Chen, Baijun Chen, Zijian Cai, Yibin Liu, Zixuan Li,
  Qiwei Liang, Xianliang Lin, Yiheng Ge, Zhenyu Gu, et~al.
\newblock {RoboTwin 2.0: A Scalable Data Generator and Benchmark with Strong
  Domain Randomization for Robust Bimanual Robotic Manipulation}.
\newblock {\em arXiv preprint arXiv:2506.18088}, 2025.

\bibitem{shang2026worldarena2}
Yu~Shang, Yinzhou Tang, Yiding Ma, Zhuohang Li, Lei Jin, Weikang Su, Xin Jin,
  Zhaolu Wang, Ziyou Wang, Xin Zhang, et~al.
\newblock {WorldArena 2.0: Extending Embodied World Model Benchmarking on
  Modality, Functionality and Platform}.
\newblock {\em arXiv preprint arXiv:2605.17912}, 2026.

\bibitem{yue2025ewmbench}
Hu~Yue, Siyuan Huang, Yue Liao, Shengcong Chen, Pengfei Zhou, Liliang Chen,
  Maoqing Yao, and Guanghui Ren.
\newblock {EWMBench: Evaluating Scene, Motion, and Semantic Quality in Embodied
  World Models}.
\newblock {\em arXiv preprint arXiv:2505.09694}, 2025.

\bibitem{pmlr-v267-qin25f}
Yiran Qin, Zhelun Shi, Jiwen Yu, Xijun Wang, Enshen Zhou, Lijun Li, Zhenfei
  Yin, Xihui Liu, Lu~Sheng, Jing Shao, Lei Bai, and Ruimao Zhang.
\newblock {WorldSimBench: Towards Video Generation Models as World Simulators}.
\newblock In Aarti Singh, Maryam Fazel, Daniel Hsu, Simon Lacoste-Julien, Felix
  Berkenkamp, Tegan Maharaj, Kiri Wagstaff, and Jerry Zhu, editors, {\em
  Proceedings of the 42nd International Conference on Machine Learning}, volume
  267 of {\em Proceedings of Machine Learning Research}, pages 50338--50362.
  PMLR, 13--19 Jul 2025.

\bibitem{jiang2026robowmbench}
Feng Jiang, Yang Chen, Kyle Xu, Yuchen Liu, Haifeng Wang, Zhenhao Shen, Jasper
  Lu, Shengze Huang, Yuanfei Wang, Chen Xie, and Ruihai Wu.
\newblock {RoboWM-Bench: A Benchmark for Evaluating World Models in Robotic
  Manipulation}.
\newblock In {\em Proceedings of the IEEE/CVF Conference on Computer Vision and
  Pattern Recognition (CVPR) Workshops}, June 2026.

\bibitem{co2026worldsimprobe}
Peterson Co, Sicheng Hu, Chunxuan Jiao, Hongyang Cheng, Yulin Luo, Yijie Xu,
  Sixiang Chen, Zhongxia Zhao, Zihao Wang, DaFeng Chi, Peidong Liu, YuTong
  Chen, Henghua Liu, Zhihao Yuan, Huizhu Jia, Yuzheng Zhuang, Tianle Zhang,
  Liang Lin, Huajie Tan, and Shanghang Zhang.
\newblock {WorldSimProbe: Diagnosing Simulator Faithfulness in
  Action-Conditioned World Models for Embodied Manipulation}, 2026.

\bibitem{liu2026kinebench}
Zeyu Liu, Zhangzhe Zhu, Yang Zhang, Chenyou Fan, Chenjia Bai, and Xuelong Li.
\newblock {KineBench: Benchmarking Embodied World Models via IDM-Free Kinematic
  Grounding}, 2026.

\bibitem{chen2026worldecho}
Sixiang Chen, Jiaming Liu, Jixian Wu, Yichen Guo, Tinghao Wang, Siyuan Qian,
  Hao Chen, Jiajun Cao, Jian Tang, and Shanghang Zhang.
\newblock {Do Robotic World Models Really Follow Actions? Diagnosing and
  Aligning Action-Conditioned Generation for Policy Learning}, 2026.

\bibitem{wang2025roboeval}
Yi~Ru Wang, Carter Ung, Christopher Tan, Grant Tannert, Jiafei Duan, Josephine
  Li, Anh Le, Rishabh Oswal, Markus Grotz, Wilbert Pumacay, et~al.
\newblock {RoboEval: Where Robotic Manipulation Meets Structured and Scalable
  Evaluation}.
\newblock {\em arXiv preprint arXiv:2507.00435}, 2025.

\bibitem{yang2026mirabench}
Tianzhuo Yang, Zihan Shen, Zirui Mi, Zhaoyi Zhang, Jiayi Zhou, Jiaming Ji,
  Juntao Dai, Jiawei Chen, Boyuan Chen, and Yaodong Yang.
\newblock {MiraBench: Evaluating Action-Conditioned Reliability in Robotic
  World Models}.
\newblock {\em arXiv preprint arXiv:2605.29360}, 2026.

\bibitem{zhang2026survey}
Zheng Zhang, Zhixiang Guo, Yuan Si, Siyuan Liang, Shunyu Liu, Weibo Gao, Song
  Wang, Leszek Rutkowski, Giuseppe Valenzise, Ming-Hsuan Yang, Lin~William
  Cong, M.~Jamal Deen, Sally Cripps, and Dacheng Tao.
\newblock {A Survey of World Model Benchmarks}.
\newblock Preprints.org, 2026.

\bibitem{liu2023reflect}
Zeyi Liu, Arpit Bahety, and Shuran Song.
\newblock {{REFLECT}: Summarizing Robot Experiences for Failure Explanation and
  Correction}.
\newblock In {\em Proceedings of The 7th Conference on Robot Learning}, volume
  229 of {\em Proceedings of Machine Learning Research}, pages 3468--3484.
  PMLR, 2023.

\bibitem{ICLR2025_70a06501}
Jiafei Duan, Wilbert Pumacay, Nishanth Kumar, Yi~Ru Wang, Shulin Tian, Wentao
  Yuan, Ranjay Krishna, Dieter Fox, Ajay Mandlekar, and Yijie Guo.
\newblock {AHA: A Vision-Language-Model for Detecting and Reasoning Over
  Failures in Robotic Manipulation}.
\newblock In Y.~Yue, A.~Garg, N.~Peng, F.~Sha, and R.~Yu, editors, {\em
  International Conference on Learning Representations}, volume 2025, pages
  45493--45517, 2025.

\bibitem{pmlr-v270-agia25a}
Christopher Agia, Rohan Sinha, Jingyun Yang, Ziang Cao, Rika Antonova, Marco
  Pavone, and Jeannette Bohg.
\newblock {Unpacking Failure Modes of Generative Policies: Runtime Monitoring
  of Consistency and Progress}.
\newblock In Pulkit Agrawal, Oliver Kroemer, and Wolfram Burgard, editors, {\em
  Proceedings of The 8th Conference on Robot Learning}, volume 270 of {\em
  Proceedings of Machine Learning Research}, pages 689--723. PMLR, 06--09 Nov
  2025.

\bibitem{lee2026roboreward}
Tony Lee, Andrew Wagenmaker, Karl Pertsch, Percy Liang, Sergey Levine, and
  Chelsea Finn.
\newblock {RoboReward: General-Purpose Vision-Language Reward Models for
  Robotics}.
\newblock {\em arXiv preprint arXiv:2601.00675}, 2026.

\bibitem{navasardyan2026failbench}
Zaruhi Navasardyan, Tatul Danielyan, and Hrant Davtyan.
\newblock {FailBench: How Reliable are VLMs at Judging Robot Task Success?}
\newblock {\em arXiv preprint arXiv:2609.03611}, 2026.

\bibitem{pacaud2025guardian}
Paul Pacaud, Ricardo Garcia, Shizhe Chen, and Cordelia Schmid.
\newblock {Guardian: Detecting Robotic Planning and Execution Errors with
  Vision-Language Models}.
\newblock {\em arXiv preprint arXiv:2512.01946}, 2025.

\bibitem{intelligence2025pi05visionlanguageactionmodelopenworld}
Physical Intelligence, Kevin Black, Noah Brown, James Darpinian, Karan
  Dhabalia, Danny Driess, Adnan Esmail, Michael Equi, Chelsea Finn, Niccolo
  Fusai, Manuel~Y. Galliker, Dibya Ghosh, Lachy Groom, Karol Hausman, Brian
  Ichter, Szymon Jakubczak, Tim Jones, Liyiming Ke, Devin LeBlanc, Sergey
  Levine, Adrian Li-Bell, Mohith Mothukuri, Suraj Nair, Karl Pertsch, Allen~Z.
  Ren, Lucy~Xiaoyang Shi, Laura Smith, Jost~Tobias Springenberg, Kyle
  Stachowicz, James Tanner, Quan Vuong, Homer Walke, Anna Walling, Haohuan
  Wang, Lili Yu, and Ury Zhilinsky.
\newblock {$\pi_{0.5}$: a Vision-Language-Action Model with Open-World
  Generalization}, 2025.

\bibitem{wu2026foundation}
Wei Wu, Fangjing Wang, Fan Lu, He~Sun, Shi Liu, Yunnan Wang, Yibin Yan, Yong
  Wang, Shuailei Ma, Xinyang Wang, et~al.
\newblock {From Foundation to Application: Improving VLA Models in Practice}.
\newblock {\em arXiv preprint arXiv:2607.06403}, 2026.

\bibitem{yuan2026fast}
Tianyuan Yuan, Zibin Dong, Yicheng Liu, and Hang Zhao.
\newblock {Fast-WAM: Do World Action Models Need Test-time Future Imagination?}
\newblock {\em arXiv preprint arXiv:2603.16666}, 2026.

\bibitem{team2026kairos}
Kairos Team, Fei Wang, Shan You, Qiming Zhang, Tao Huang, Zuoyi Fu, Zhisheng
  Zheng, Yunlong Xi, Feng Lv, Xiaoming Wu, et~al.
\newblock {Kairos: A Regret-Aware Native World-Action Model Stack for Physical
  AI}.
\newblock {\em arXiv preprint arXiv:2606.16533}, 2026.

\bibitem{ali2025world}
Arslan Ali, Junjie Bai, Maciej Bala, Yogesh Balaji, Aaron Blakeman, Tiffany
  Cai, Jiaxin Cao, Tianshi Cao, Elizabeth Cha, Yu-Wei Chao, et~al.
\newblock {World Simulation with Video Foundation Models for Physical AI}.
\newblock {\em arXiv preprint arXiv:2511.00062}, 2025.

\bibitem{bwm2026bwm}
{BWM Team}.
\newblock {BWM: A Low-Cost High-Fidelity World Simulator for Robot Learning}.
\newblock {\em arXiv preprint arXiv:2607.29302}, 2026.

\bibitem{gao2026dreamdojo}
Shenyuan Gao, William Liang, Kaiyuan Zheng, Ayaan Malik, Seonghyeon Ye, Sihyun
  Yu, Wei-Cheng Tseng, Yuzhu Dong, Kaichun Mo, Chen-Hsuan Lin, Qianli Ma,
  Seungjun Nah, Loic Magne, Jiannan Xiang, Yuqi Xie, Ruijie Zheng, Dantong Niu,
  You~Liang Tan, K.R. Zentner, George Kurian, Suneel Indupuru, Pooya Jannaty,
  Jinwei Gu, Jun Zhang, Jitendra Malik, Pieter Abbeel, Ming-Yu Liu, Yuke Zhu,
  Joel Jang, and Linxi Fan.
\newblock {DreamDojo: A Generalist Robot World Model from Large-Scale Human
  Videos}.
\newblock In {\em International Conference on Machine Learning (ICML)}, 2026.

\bibitem{qiu2026ge}
Boxiang Qiu, Liliang Chen, Yue Liao, Nan Wang, Lintao Wang, Jiayi Luo, Wenzhi
  Zhao, Shengcong Chen, Di~Chen, Ye~Li, et~al.
\newblock {GE-Sim 2.0: A Roadmap Towards Comprehensive Closed-loop Video World
  Simulators for Robotic Manipulation}.
\newblock {\em arXiv preprint arXiv:2605.27491}, 2026.

\bibitem{huang2026learningtransferabledynamicspriors}
Ze~Huang, Jiahui Zhang, Hairuo Liu, Chenxi Zhang, Ran Cheng, and Li~Zhang.
\newblock {Learning Transferable Dynamics Priors from Action to World
  Modeling}, 2026.

\bibitem{dexmal2026dw05}
{Dexmal}.
\newblock {{DW05}: A Multimodal World Model for Embodied Intelligence}.
\newblock GitHub repository, 2026.

\end{thebibliography}
\clearpage
\appendix
\section*{\titlefont\sffamily\themeblue{Appendix}}
\phantomsection
\addcontentsline{toc}{section}{Appendix}
\makeatletter
\setlength{\@fptop}{0pt}
\setlength{\@fpsep}{16pt}
\setlength{\@fpbot}{0pt plus 1fil}
\makeatother
\section{Tasks and Experimental Settings}
\label{app:experimental_details}
\subsection{Tasks, Success Criteria, and Stages}
\label{app:task_details}
\label{app:stage_decomposition}
Failure-reason and failure-stage evaluation covers 20 RoboTwin tasks, all ten LIBERO-10 tasks, and two real-world tasks on G1. For each platform, the task descriptions and success criteria are followed by the stage definitions shared across policies and world models.

Each task--policy setting contains 50 reference episodes. Reference execution stops upon success or exhaustion of its action budget.

\subsubsection{RoboTwin}
RoboTwin spans bimanual coordination, spatial arrangement, tool use, articulated-object interaction, and sequential manipulation. Table~\ref{tab:app_robotwin_tasks} lists task instructions, native success criteria, and limits on waypoint commands, each potentially spanning multiple physics steps.
\begingroup
\fontsize{8.5}{10.2}\selectfont
\setlength{\tabcolsep}{3pt}
\renewcommand{\arraystretch}{1.06}
\setlength{\LTpre}{5pt}
\setlength{\LTpost}{8pt}
\begin{longtable}{@{}>{\raggedright\arraybackslash}p{\dimexpr0.240\linewidth-1.5\tabcolsep\relax}>{\raggedright\arraybackslash}p{\dimexpr0.250\linewidth-1.5\tabcolsep\relax}>{\raggedright\arraybackslash}p{\dimexpr0.430\linewidth-1.5\tabcolsep\relax}>{\raggedright\arraybackslash}p{\dimexpr0.080\linewidth-1.5\tabcolsep\relax}@{}}
\caption{RoboTwin tasks for failure-reason and failure-stage evaluation under \texttt{demo\_clean}. Instructions are abbreviated; policies receive the original episode-specific text. Success conditions summarize native checks. Limits count waypoint commands.}\label{tab:app_robotwin_tasks}\\
\toprule
\textbf{Task / capability} & \textbf{Instruction} & \textbf{Native success criterion} & \textbf{Limit}\\
\midrule
\endfirsthead
\multicolumn{4}{@{}l}{\tablename~\thetable{} (continued)}\\[3pt]
\toprule
\textbf{Task / capability} & \textbf{Instruction} & \textbf{Native success criterion} & \textbf{Limit}\\
\midrule
\endhead
\midrule
\multicolumn{4}{r@{}}{\footnotesize Continued on next page}\\
\endfoot
\bottomrule
\endlastfoot
\texttt{beat\_\allowbreak{}block\_\allowbreak{}hammer}\par\smallskip\textit{Tool use} & Pick up the hammer and strike the block. & Hammer contacts the block, with planar functional-point errors below 0.02 m per axis. & 400\\
\addlinespace[3pt]
\texttt{blocks\_\allowbreak{}ranking\_\allowbreak{}rgb}\par\smallskip\textit{Color-based arrangement} & Arrange the blocks in red--green--blue order. & Correct left-to-right order and adjacent spacing; both grippers open. & 1200\\
\addlinespace[3pt]
\texttt{blocks\_\allowbreak{}ranking\_\allowbreak{}size}\par\smallskip\textit{Size-based arrangement} & Arrange the blocks from large to small. & Correct left-to-right size order and adjacent spacing; both grippers open. & 1200\\
\addlinespace[3pt]
\texttt{handover\_\allowbreak{}block}\par\smallskip\textit{Bimanual transfer} & Transfer the block between arms and place it at the target. & Block reaches the target within position tolerances; right gripper open. & 800\\
\addlinespace[3pt]
\texttt{hanging\_\allowbreak{}mug}\par\smallskip\textit{Reorientation and hanging} & Reorient the mug and hang it on the rack. & Mug functional point aligns with the rack target and satisfies the height threshold; right gripper open. & 900\\
\addlinespace[3pt]
\texttt{move\_\allowbreak{}pillbottle\_\allowbreak{}pad}\par\smallskip\textit{Precise placement} & Place the pill bottle on the pad. & Bottle satisfies planar alignment and resting-height tolerances; both grippers open. & 400\\
\addlinespace[3pt]
\texttt{open\_\allowbreak{}laptop}\par\smallskip\textit{Articulated-object interaction} & Open the laptop. & Hinge reaches 40\% of its joint range; acting end effector remains near the designated contact point. & 700\\
\addlinespace[3pt]
\texttt{pick\_\allowbreak{}diverse\_\allowbreak{}bottles}\par\smallskip\textit{Bimanual grasping} & Pick up one bottle with each arm. & Both bottles reach their respective planar target regions and required heights. & 400\\
\addlinespace[3pt]
\texttt{place\_\allowbreak{}a2b\_\allowbreak{}left}\par\smallskip\textit{Spatial positioning} & Place object A to the left of object B. & Correct side, planar distance of 0.08--0.20 m, and lateral alignment; both grippers open. & 400\\
\addlinespace[3pt]
\texttt{place\_\allowbreak{}bread\_\allowbreak{}skillet}\par\smallskip\textit{Coordinated holding and placement} & Hold the skillet and place the bread inside. & Bread and skillet functional point satisfy planar alignment and height constraints. & 500\\
\addlinespace[3pt]
\texttt{place\_\allowbreak{}cans\_\allowbreak{}plasticbox}\par\smallskip\textit{Multi-object placement} & Place both cans in the plastic box. & Each can is within 0.04 m planar distance of a box target point; both grippers open. & 800\\
\addlinespace[3pt]
\texttt{place\_\allowbreak{}dual\_\allowbreak{}shoes}\par\smallskip\textit{Oriented placement} & Place both shoes in the box, pointing left. & Both shoes satisfy target position, height, and orientation constraints; both grippers open. & 600\\
\addlinespace[3pt]
\texttt{place\_\allowbreak{}empty\_\allowbreak{}cup}\par\smallskip\textit{Precise placement} & Place the cup on the coaster. & Functional-point planar distance below 0.035 m and vertical error below 0.015 m; both grippers open. & 500\\
\addlinespace[3pt]
\texttt{place\_\allowbreak{}object\_\allowbreak{}scale}\par\smallskip\textit{Placement on a support surface} & Place the object on the scale. & Object satisfies planar alignment and height constraints; acting gripper open. & 400\\
\addlinespace[3pt]
\texttt{place\_\allowbreak{}phone\_\allowbreak{}stand}\par\smallskip\textit{Three-dimensional alignment} & Place the phone on its stand. & Phone and stand functional points satisfy position tolerances; both grippers open. & 400\\
\addlinespace[3pt]
\texttt{put\_\allowbreak{}bottles\_\allowbreak{}dustbin}\par\smallskip\textit{Sequential manipulation} & Put all three bottles in the dustbin. & All bottles lie within the specified planar region and height interval. & 1700\\
\addlinespace[3pt]
\texttt{put\_\allowbreak{}object\_\allowbreak{}cabinet}\par\smallskip\textit{Drawer interaction and placement} & Open the drawer and place the object inside. & Object reaches the cabinet target region and height range; acting gripper open. & 700\\
\addlinespace[3pt]
\texttt{scan\_\allowbreak{}object}\par\smallskip\textit{Bimanual tool alignment} & Hold the object and align the scanner with it. & Scanner-axis alignment and distance constraints hold; both grippers closed. & 500\\
\addlinespace[3pt]
\texttt{stack\_\allowbreak{}blocks\_\allowbreak{}three}\par\smallskip\textit{Sequential stacking} & Stack the three blocks in order. & Consecutive blocks satisfy the required vertical offsets and position tolerances; both grippers open. & 1200\\
\addlinespace[3pt]
\texttt{stack\_\allowbreak{}bowls\_\allowbreak{}three}\par\smallskip\textit{Multi-object stacking} & Stack the three bowls. & Height-sorted bowls satisfy adjacent planar alignment and native one-sided height bounds; both grippers open. & 1200\\
\end{longtable}
\endgroup

Success follows the implemented predicates: \texttt{handover\_block} checks terminal placement rather than the transfer sequence, and \texttt{put\_object\_cabinet} does not require drawer closure.

Table~\ref{tab:app_robotwin_stages} lists task-specific stage categories. Both \texttt{blocks\_ranking\_rgb} and \texttt{blocks\_ranking\_size} use eight categories: repeated Reach and Place actions share labels, while grasps distinguish the target blocks and transfers distinguish their directions. Slash-separated parallel actions form one stage. Names are translated into English for presentation; annotation retains the source labels.
\begingroup
\fontsize{9}{10.8}\selectfont
\setlength{\tabcolsep}{3pt}
\renewcommand{\arraystretch}{1.06}
\setlength{\LTpre}{5pt}
\setlength{\LTpost}{8pt}
\begin{longtable}{@{}>{\raggedright\arraybackslash}p{\dimexpr0.270\linewidth-1\tabcolsep\relax}>{\raggedright\arraybackslash}p{\dimexpr0.730\linewidth-1\tabcolsep\relax}@{}}
\caption{Task-specific RoboTwin stage categories. Repeated Reach and Place actions share labels in \texttt{blocks\_ranking\_rgb} and \texttt{blocks\_ranking\_size}; other repeated actions retain separate sequence positions. Slash-separated parallel actions form one stage.}\label{tab:app_robotwin_stages}\\
\toprule
\textbf{Task} & \textbf{Stages}\\
\midrule
\endfirsthead
\multicolumn{2}{@{}l}{\tablename~\thetable{} (continued)}\\[3pt]
\toprule
\textbf{Task} & \textbf{Stages}\\
\midrule
\endhead
\midrule
\multicolumn{2}{r@{}}{\footnotesize Continued on next page}\\
\endfoot
\bottomrule
\endlastfoot
\texttt{beat\_\allowbreak{}block\_\allowbreak{}hammer} & \textbf{1.}~Reach; \textbf{2.}~Grasp; \textbf{3.}~Transfer; \textbf{4.}~Strike\\
\addlinespace[3pt]
\texttt{blocks\_\allowbreak{}ranking\_\allowbreak{}rgb} & \textbf{1.}~Reach; \textbf{2.}~Grasp red; \textbf{3.}~Transfer right; \textbf{4.}~Place; \textbf{5.}~Grasp green; \textbf{6.}~Transfer center; \textbf{7.}~Grasp blue; \textbf{8.}~Transfer left\\
\addlinespace[3pt]
\texttt{blocks\_\allowbreak{}ranking\_\allowbreak{}size} & \textbf{1.}~Reach; \textbf{2.}~Grasp small; \textbf{3.}~Transfer right; \textbf{4.}~Place; \textbf{5.}~Grasp medium; \textbf{6.}~Transfer center; \textbf{7.}~Grasp large; \textbf{8.}~Transfer left\\
\addlinespace[3pt]
\texttt{handover\_\allowbreak{}block} & \textbf{1.}~Arm 1: Reach; \textbf{2.}~Arm 1: Grasp; \textbf{3.}~Arm 1: Transfer; \textbf{4.}~Arm 2: Reach; \textbf{5.}~Arm 2: Grasp; \textbf{6.}~Arm 1: Release; \textbf{7.}~Arm 2: Transfer; \textbf{8.}~Arm 2: Place\\
\addlinespace[3pt]
\texttt{hanging\_\allowbreak{}mug} & \textbf{1.}~Arm 1: Reach; \textbf{2.}~Arm 1: Grasp; \textbf{3.}~Arm 1: Place; \textbf{4.}~Arm 2: Reach; \textbf{5.}~Arm 2: Grasp; \textbf{6.}~Arm 2: Transfer; \textbf{7.}~Arm 2: Place\\
\addlinespace[3pt]
\texttt{move\_\allowbreak{}pillbottle\_\allowbreak{}pad} & \textbf{1.}~Reach; \textbf{2.}~Grasp; \textbf{3.}~Transfer; \textbf{4.}~Place\\
\addlinespace[3pt]
\texttt{open\_\allowbreak{}laptop} & \textbf{1.}~Reach; \textbf{2.}~Grasp; \textbf{3.}~Lift\\
\addlinespace[3pt]
\texttt{pick\_\allowbreak{}diverse\_\allowbreak{}bottles} & \textbf{1.}~Arm 1: Reach / Arm 2: Reach; \textbf{2.}~Arm 1: Grasp / Arm 2: Grasp; \textbf{3.}~Arm 1: Transfer / Arm 2: Transfer\\
\addlinespace[3pt]
\texttt{place\_\allowbreak{}a2b\_\allowbreak{}left} & \textbf{1.}~Reach; \textbf{2.}~Grasp A; \textbf{3.}~Place A\\
\addlinespace[3pt]
\texttt{place\_\allowbreak{}bread\_\allowbreak{}skillet} & \textbf{1.}~Arm 1: Reach / Arm 2: Reach; \textbf{2.}~Arm 1: Grasp / Arm 2: Grasp; \textbf{3.}~Rotate the skillet; \textbf{4.}~Transfer the bread; \textbf{5.}~Place the bread\\
\addlinespace[3pt]
\texttt{place\_\allowbreak{}cans\_\allowbreak{}plasticbox} & \textbf{1.}~Arm 1: Reach / Arm 2: Reach; \textbf{2.}~Arm 1: Grasp / Arm 2: Grasp; \textbf{3.}~Arm 1: Transfer; \textbf{4.}~Arm 1: Place; \textbf{5.}~Arm 1: Retract; \textbf{6.}~Arm 2: Transfer; \textbf{7.}~Arm 2: Place; \textbf{8.}~Arm 2: Retract\\
\addlinespace[3pt]
\texttt{place\_\allowbreak{}dual\_\allowbreak{}shoes} & \textbf{1.}~Arm 1: Reach / Arm 2: Reach; \textbf{2.}~Arm 1: Grasp / Arm 2: Grasp; \textbf{3.}~Arm 1: Transfer; \textbf{4.}~Arm 1: Place; \textbf{5.}~Arm 1: Retract; \textbf{6.}~Arm 2: Transfer; \textbf{7.}~Arm 2: Place\\
\addlinespace[3pt]
\texttt{place\_\allowbreak{}empty\_\allowbreak{}cup} & \textbf{1.}~Reach; \textbf{2.}~Grasp; \textbf{3.}~Transfer; \textbf{4.}~Place\\
\addlinespace[3pt]
\texttt{place\_\allowbreak{}object\_\allowbreak{}scale} & \textbf{1.}~Reach; \textbf{2.}~Grasp; \textbf{3.}~Transfer; \textbf{4.}~Place\\
\addlinespace[3pt]
\texttt{place\_\allowbreak{}phone\_\allowbreak{}stand} & \textbf{1.}~Reach; \textbf{2.}~Grasp; \textbf{3.}~Transfer; \textbf{4.}~Place\\
\addlinespace[3pt]
\texttt{put\_\allowbreak{}bottles\_\allowbreak{}dustbin} & \textbf{1.}~Bottle 1: Reach; \textbf{2.}~Bottle 1: Grasp; \textbf{3.}~Bottle 1: Transfer; \textbf{4.}~Bottle 1: Place; \textbf{5.}~Bottle 2: Reach; \textbf{6.}~Bottle 2: Grasp; \textbf{7.}~Bottle 2: Transfer; \textbf{8.}~Bottle 2: Place; \textbf{9.}~Bottle 3: Reach; \textbf{10.}~Bottle 3: Grasp; \textbf{11.}~Bottle 3: Transfer; \textbf{12.}~Bottle 3: Place\\
\addlinespace[3pt]
\texttt{put\_\allowbreak{}object\_\allowbreak{}cabinet} & \textbf{1.}~Arm 1: Reach; \textbf{2.}~Arm 1: Grasp; \textbf{3.}~Arm 2: Reach; \textbf{4.}~Arm 2: Grasp; \textbf{5.}~Arm 2: Pull open; \textbf{6.}~Arm 1: Transfer; \textbf{7.}~Arm 1: Place\\
\addlinespace[3pt]
\texttt{scan\_\allowbreak{}object} & \textbf{1.}~Arms 1 and 2: Reach; \textbf{2.}~Arms 1 and 2: Grasp; \textbf{3.}~Arm 1: Transfer; \textbf{4.}~Arm 2: Transfer\\
\addlinespace[3pt]
\texttt{stack\_\allowbreak{}blocks\_\allowbreak{}three} & \textbf{1.}~Block 1: Reach; \textbf{2.}~Block 1: Grasp; \textbf{3.}~Block 1: Place; \textbf{4.}~Block 1: Retract; \textbf{5.}~Block 2: Reach; \textbf{6.}~Block 2: Grasp; \textbf{7.}~Block 2: Place; \textbf{8.}~Block 2: Retract; \textbf{9.}~Block 3: Reach; \textbf{10.}~Block 3: Grasp; \textbf{11.}~Block 3: Place; \textbf{12.}~Block 3: Retract\\
\addlinespace[3pt]
\texttt{stack\_\allowbreak{}bowls\_\allowbreak{}three} & \textbf{1.}~Bowl 1: Reach; \textbf{2.}~Bowl 1: Grasp; \textbf{3.}~Bowl 1: Place; \textbf{4.}~Bowl 1: Retract; \textbf{5.}~Bowl 2: Reach; \textbf{6.}~Bowl 2: Grasp; \textbf{7.}~Bowl 2: Place; \textbf{8.}~Bowl 2: Retract; \textbf{9.}~Bowl 3: Reach; \textbf{10.}~Bowl 3: Grasp; \textbf{11.}~Bowl 3: Place; \textbf{12.}~Bowl 3: Retract\\
\end{longtable}
\endgroup

\subsubsection{LIBERO-10}
LIBERO-10 contains long-horizon, multi-object tasks with composed goals (Table~\ref{tab:app_libero_tasks}). Three $\pi_{0.5}$ data variants give 150 reference episodes per task and 1,500 in total. The $\pi_{0.5}$ variants use 520 policy actions at 20\,Hz, after ten settling steps excluded from the budget.
\begingroup
\fontsize{8.5}{10.2}\selectfont
\setlength{\tabcolsep}{3pt}
\renewcommand{\arraystretch}{1.06}
\setlength{\LTpre}{5pt}
\setlength{\LTpost}{8pt}
\begin{longtable}{@{}>{\raggedright\arraybackslash}p{\dimexpr0.240\linewidth-1.5\tabcolsep\relax}>{\raggedright\arraybackslash}p{\dimexpr0.250\linewidth-1.5\tabcolsep\relax}>{\raggedright\arraybackslash}p{\dimexpr0.430\linewidth-1.5\tabcolsep\relax}>{\raggedright\arraybackslash}p{\dimexpr0.080\linewidth-1.5\tabcolsep\relax}@{}}
\caption{Complete LIBERO-10 task set. All listed goal predicates must hold simultaneously. Limits count policy actions and exclude the ten settling steps.}\label{tab:app_libero_tasks}\\
\toprule
\textbf{Task / capability} & \textbf{Instruction} & \textbf{Native success criterion} & \textbf{Limit}\\
\midrule
\endfirsthead
\multicolumn{4}{@{}l}{\tablename~\thetable{} (continued)}\\[3pt]
\toprule
\textbf{Task / capability} & \textbf{Instruction} & \textbf{Native success criterion} & \textbf{Limit}\\
\midrule
\endhead
\midrule
\multicolumn{4}{r@{}}{\footnotesize Continued on next page}\\
\endfoot
\bottomrule
\endlastfoot
\texttt{task\_\allowbreak{}00}\par\smallskip\textit{Two-object placement} & Put both the alphabet soup and the tomato sauce in the basket. & Both objects are inside the basket. & 520\\
\addlinespace[3pt]
\texttt{task\_\allowbreak{}01}\par\smallskip\textit{Two-object placement} & Put both the cream cheese box and the butter in the basket. & Both objects are inside the basket. & 520\\
\addlinespace[3pt]
\texttt{task\_\allowbreak{}02}\par\smallskip\textit{State change and placement} & Turn on the stove and put the moka pot on it. & Stove is on; moka pot is on its cooking region. & 520\\
\addlinespace[3pt]
\texttt{task\_\allowbreak{}03}\par\smallskip\textit{Placement and drawer closure} & Put the black bowl in the bottom drawer of the cabinet and close it. & Bowl is inside the bottom drawer; drawer is closed. & 520\\
\addlinespace[3pt]
\texttt{task\_\allowbreak{}04}\par\smallskip\textit{Object--target association} & Put the white mug on the left plate and put the yellow and white mug on the right plate. & Each mug is on its designated plate. & 520\\
\addlinespace[3pt]
\texttt{task\_\allowbreak{}05}\par\smallskip\textit{Compartment placement} & Pick up the book and place it in the back compartment of the caddy. & Book is inside the caddy's rear compartment. & 520\\
\addlinespace[3pt]
\texttt{task\_\allowbreak{}06}\par\smallskip\textit{Composed spatial goals} & Put the white mug on the plate and put the chocolate pudding to the right of the plate. & Mug is on the plate; pudding is in the designated region to its right. & 520\\
\addlinespace[3pt]
\texttt{task\_\allowbreak{}07}\par\smallskip\textit{Two-object placement} & Put both the alphabet soup and the cream cheese box in the basket. & Both objects are inside the basket. & 520\\
\addlinespace[3pt]
\texttt{task\_\allowbreak{}08}\par\smallskip\textit{Multi-object placement} & Put both moka pots on the stove. & Both pots are on the cooking region; stove remains on. & 520\\
\addlinespace[3pt]
\texttt{task\_\allowbreak{}09}\par\smallskip\textit{Placement and door closure} & Put the yellow and white mug in the microwave and close it. & Mug is inside the microwave's heating region; door is closed. & 520\\
\end{longtable}
\endgroup

Success uses the native BDDL predicates \texttt{In}, \texttt{On}, \texttt{Close}, and \texttt{Turnon}. For \texttt{task\_08}, the stove is already on at initialization. Table~\ref{tab:app_libero_stages} lists the corresponding subgoal sequences.
\begingroup
\fontsize{9}{10.8}\selectfont
\setlength{\tabcolsep}{3pt}
\renewcommand{\arraystretch}{1.06}
\setlength{\LTpre}{5pt}
\setlength{\LTpost}{8pt}
\begin{longtable}{@{}>{\raggedright\arraybackslash}p{\dimexpr0.130\linewidth-1\tabcolsep\relax}>{\raggedright\arraybackslash}p{\dimexpr0.870\linewidth-1\tabcolsep\relax}@{}}
\caption{Ordered stages for all LIBERO-10 tasks.}\label{tab:app_libero_stages}\\
\toprule
\textbf{Task} & \textbf{Ordered stages}\\
\midrule
\endfirsthead
\multicolumn{2}{@{}l}{\tablename~\thetable{} (continued)}\\[3pt]
\toprule
\textbf{Task} & \textbf{Ordered stages}\\
\midrule
\endhead
\midrule
\multicolumn{2}{r@{}}{\footnotesize Continued on next page}\\
\endfoot
\bottomrule
\endlastfoot
\texttt{task\_\allowbreak{}00} & \textbf{1.}~Grasp the alphabet soup.; \textbf{2.}~Move the alphabet soup to the basket.; \textbf{3.}~Release the alphabet soup and retreat the arm.; \textbf{4.}~Grasp the tomato sauce.; \textbf{5.}~Move the tomato sauce to the basket.; \textbf{6.}~Release the tomato sauce and retreat the arm.\\
\addlinespace[3pt]
\texttt{task\_\allowbreak{}01} & \textbf{1.}~Grasp the cream cheese box.; \textbf{2.}~Move the cream cheese box to the basket.; \textbf{3.}~Release the cream cheese box and retreat the arm.; \textbf{4.}~Grasp the butter.; \textbf{5.}~Move the butter to the basket.; \textbf{6.}~Release the butter and retreat the arm.\\
\addlinespace[3pt]
\texttt{task\_\allowbreak{}02} & \textbf{1.}~Turn on the stove.; \textbf{2.}~Grasp the moka pot.; \textbf{3.}~Move the moka pot to the stove.; \textbf{4.}~Place the moka pot on the stove, release it, and retreat the arm.\\
\addlinespace[3pt]
\texttt{task\_\allowbreak{}03} & \textbf{1.}~Open the bottom drawer.; \textbf{2.}~Grasp the black bowl.; \textbf{3.}~Move the black bowl into the bottom drawer.; \textbf{4.}~Release the bowl and retreat the arm.; \textbf{5.}~Close the bottom drawer and retreat the arm.\\
\addlinespace[3pt]
\texttt{task\_\allowbreak{}04} & \textbf{1.}~Grasp the white mug.; \textbf{2.}~Move the white mug to the left plate.; \textbf{3.}~Release the white mug and retreat the arm.; \textbf{4.}~Grasp the yellow-and-white mug.; \textbf{5.}~Move the yellow-and-white mug to the right plate.; \textbf{6.}~Release the mug and retreat the arm.\\
\addlinespace[3pt]
\texttt{task\_\allowbreak{}05} & \textbf{1.}~Grasp the book.; \textbf{2.}~Move the book to the back compartment of the caddy.; \textbf{3.}~Release the book and retreat the arm.\\
\addlinespace[3pt]
\texttt{task\_\allowbreak{}06} & \textbf{1.}~Grasp the white mug.; \textbf{2.}~Move the white mug to the plate.; \textbf{3.}~Release the mug and retreat the arm.; \textbf{4.}~Grasp the chocolate pudding.; \textbf{5.}~Move the chocolate pudding to the right of the plate.; \textbf{6.}~Release the pudding and retreat the arm.\\
\addlinespace[3pt]
\texttt{task\_\allowbreak{}07} & \textbf{1.}~Grasp the alphabet soup.; \textbf{2.}~Move the alphabet soup to the basket.; \textbf{3.}~Release the soup and retreat the arm.; \textbf{4.}~Grasp the cream cheese box.; \textbf{5.}~Move the cream cheese box to the basket.; \textbf{6.}~Release the box and retreat the arm.\\
\addlinespace[3pt]
\texttt{task\_\allowbreak{}08} & \textbf{1.}~Grasp the first moka pot.; \textbf{2.}~Move the first moka pot to the stove.; \textbf{3.}~Release the first moka pot and retreat the arm.; \textbf{4.}~Grasp the second moka pot.; \textbf{5.}~Move the second moka pot to the stove.; \textbf{6.}~Release the second moka pot and retreat the arm.\\
\addlinespace[3pt]
\texttt{task\_\allowbreak{}09} & \textbf{1.}~Open the microwave door.; \textbf{2.}~Grasp the yellow-and-white mug.; \textbf{3.}~Move the mug into the microwave.; \textbf{4.}~Release the mug and retreat the arm.; \textbf{5.}~Close the microwave door and retreat the arm.\\
\end{longtable}
\endgroup

\subsubsection{Real-World Tasks on G1}
\begin{figure}[!htbp]
\centering
\includegraphics[width=\linewidth,height=0.78\textheight,keepaspectratio]{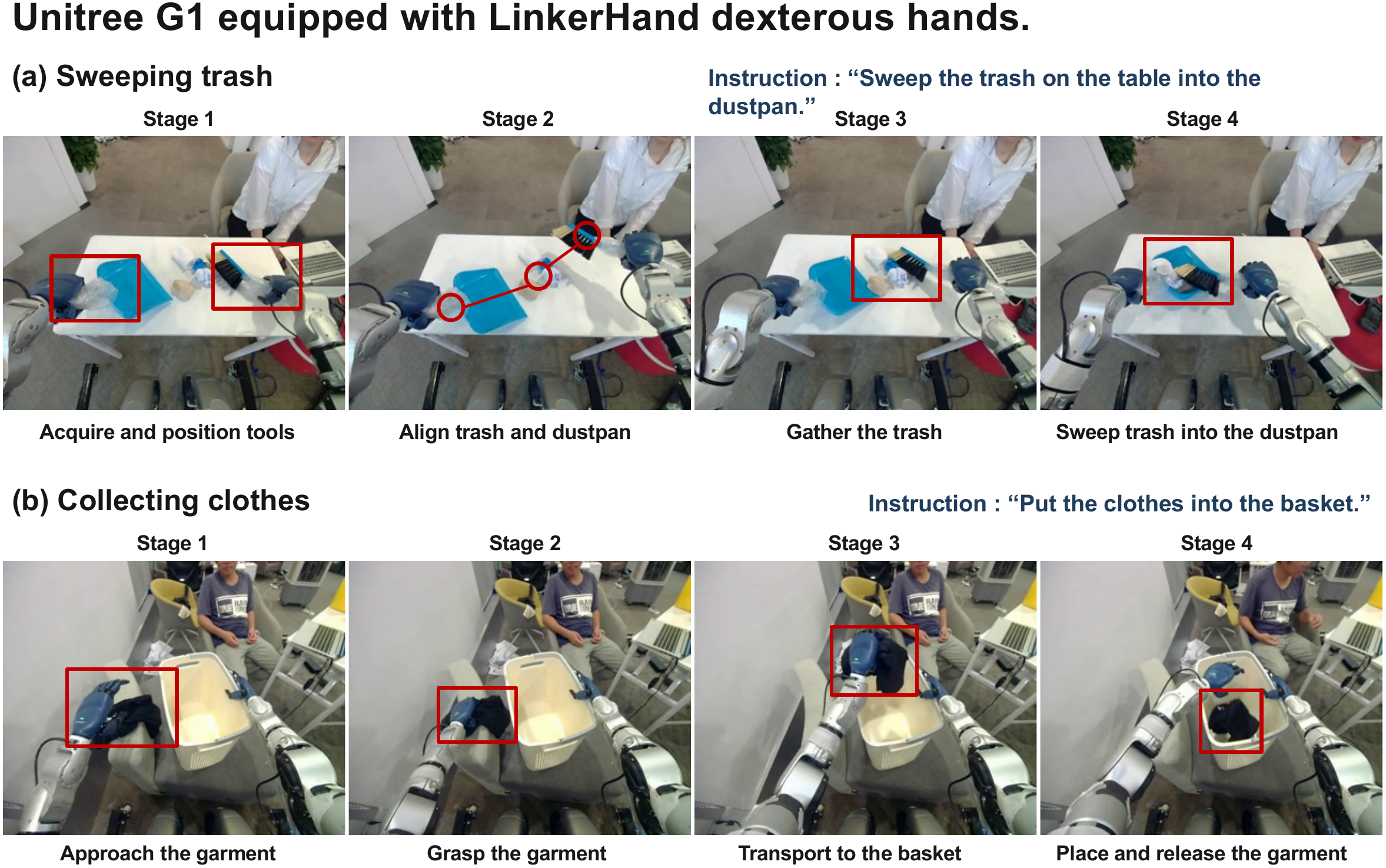}
\caption{Task scenes, instructions, and stage decompositions for two real-world tasks on G1: (a) sweeping trash and (b) collecting clothes. The robot uses LinkerHand dexterous hands; red annotations highlight relevant objects and alignment.}
\label{fig:g1_tasks_stages}
\end{figure}

Figure~\ref{fig:g1_tasks_stages} gives the scenes, language instructions, and four task-specific stages for sweeping trash and collecting clothes.

\paragraph{Sweeping trash.}
Success requires the main target trash to be swept into the dustpan using the broom. Tiny, visually ambiguous fragments may be tolerated; clearly visible larger pieces remaining outside preclude success.
\paragraph{Collecting clothes.}
Success requires the target clothing to be placed in the basket and released. Partial draping over the rim is acceptable when the basket supports the clothing; suspension above the basket or continued support by the hand does not satisfy the goal.
\FloatBarrier

\clearpage
\subsection{Environment, Model, and Policy Configurations}
\label{app:wfb:config}
\label{app:configurations}

\textbf{Evaluation scope.} WorldFailBench evaluates complete world-model systems, including their platform adapters, state-feedback rules, and execution schedules. These components vary across models and jointly determine their capabilities for closed-loop policy evaluation. We therefore follow each model's official implementation and inference procedure, adapting observation and action interfaces to the target platform where necessary.

\subsubsection{Observation and action interfaces}
\label{app:wfb:interfaces}

We evaluate robot policies in LIBERO, RoboTwin, and two real-world manipulation tasks on G1. Each policy receives a task instruction, one observed RGB view, and a platform-specific robot-state vector. When a policy interface includes additional wrist-camera slots, these slots are zero-filled or masked according to that interface. On LIBERO, all three policy variants and all world models use only the main camera view. Table~\ref{tab:wfb:interfaces} summarizes policy-facing inputs and native action representations.

\begingroup
\small
\setlength{\tabcolsep}{4pt}
\renewcommand{\arraystretch}{1.16}
\begin{longtable}{@{}>{\raggedright\arraybackslash}p{\dimexpr0.12\linewidth-2\tabcolsep\relax}>{\raggedright\arraybackslash}p{\dimexpr0.28\linewidth-2\tabcolsep\relax}>{\raggedright\arraybackslash}p{\dimexpr0.25\linewidth-2\tabcolsep\relax}>{\raggedright\arraybackslash}p{\dimexpr0.35\linewidth-2\tabcolsep\relax}@{}}
\caption{Environment and policy interfaces. Prediction horizon denotes the number of actions returned by a policy call; executed prefix or replanning interval denotes the portion advanced before another call. Action-timeline frequencies and iterative sampling steps are separate quantities. Reference commands are executed through the native platform interfaces described below.}\label{tab:wfb:interfaces}\\
\toprule
\textbf{Platform} & \textbf{Environment and policy observations} & \textbf{Policy action space} & \textbf{Policy execution settings} \\
\midrule
\endfirsthead
\multicolumn{4}{l}{\small\textit{Table~\thetable{} (continued)}}\\
\toprule
\textbf{Platform} & \textbf{Environment and policy observations} & \textbf{Policy action space} & \textbf{Policy execution settings} \\
\midrule
\endhead
\midrule
\multicolumn{4}{r}{\small\textit{Continued on next page}}\\
\endfoot
\bottomrule
\endlastfoot
LIBERO & Initial states reconstructed with MuJoCo 3.3.2 and robosuite 1.4.0; 256 \(\times\) 256 fixed front-view RGB; 8D robot state; task instruction & 7D action: end-effector translation and rotation increments plus a gripper command & The evaluated \(\pi_{0.5}\) variants predict 10 actions and execute the first 8 at 20 Hz, replanning every 0.4 s \\*
RoboTwin & \texttt{demo\_clean} scenes and \texttt{seen} instructions; head-camera RGB; 14D robot state; task instruction & 14D absolute targets: six joint targets and one gripper target per arm & 30 Hz action timeline; \(\pi_{0.5}\): prediction horizon 10; LingBot-VLA-v2: reference action chunk 50; FastWAM: reference replanning interval 24 actions and 10 sampling steps \\*
G1 & 256 \(\times\) 448 front-view RGB; 26D robot state; task instruction & 26D action & Kairos; four policy versions per task \\
\end{longtable}
\endgroup

BWM is evaluated on RoboTwin. Ctrl-World, Cosmos-Predict-2.5, DreamDojo, DW05, A2World, and GE-Sim are evaluated on all three platforms.

RoboTwin reference execution uses the native qpos-to-TOPP interface to execute joint targets. LIBERO gripper outputs are converted to the environment's native commands.

\subsubsection{Policy variants}
\label{app:wfb:policies}

Policy fine-tuning follows the official implementation. The three LIBERO policies are \(\pi_{0.5}\)-data-20, \(\pi_{0.5}\)-data-50, and the full-data \(\pi_{0.5}\). The full-data variant is named data100 in the training configuration. Their training subsets contain 305, 764, and 1,524 trajectories, respectively, from the 40-task policy-training collection, with a shared holdout of 169 trajectories. Evaluation uses the ten LIBERO-10 tasks. These trajectory counts describe policy fine-tuning.

RoboTwin evaluates the three \(\pi_{0.5}\) variants, LingBot-VLA-v2, and FastWAM. The \(\pi_{0.5}\) training collection covers 50 WorldArena tasks, with 500 trajectories for data20 and 1,250 for data50. Each subset has its own action-normalization statistics.

For each of the two G1 tasks, sweeping trash and collecting clothes, we evaluate four versions of the Kairos policy, labeled V1--V4 within each task.

Public policy implementations and checkpoint sources are available for $\pi_{0.5}$ through OpenPI (\href{https://github.com/Physical-Intelligence/openpi#fine-tuning-base-models-on-your-own-data}{fine-tuning instructions}; \href{https://github.com/Physical-Intelligence/openpi#model-checkpoints}{checkpoints}), FastWAM (\href{https://github.com/yuantianyuan01/FastWAM#training}{training code}; \href{https://huggingface.co/yuanty/fastwam}{checkpoints}), and LingBot-VLA-v2 (\href{https://github.com/Robbyant/lingbot-vla-v2#post-training-example}{fine-tuning instructions}; \href{https://huggingface.co/robbyant/lingbot-vla-v2-6b}{base checkpoint}; \href{https://huggingface.co/robbyant/lingbot-vla-v2-6b-robotwin}{RoboTwin checkpoint}). These links identify the upstream releases from which our policy variants are obtained or adapted.

\subsubsection{World-model inference and action conditioning}
\label{app:wfb:inference}

Table~\ref{tab:wfb:inference} records each model's applied image input, sampler, temporal window, and action-conditioning interface. Inference uses the official implementation and its platform adapter. Policy prediction horizon specifies an action-sequence length, video frequency specifies timestamps in the represented trajectory, and sampling iterations specify the numerical generation procedure. None of these timeline frequencies denotes wall-clock generation throughput.

\begingroup
\small
\setlength{\tabcolsep}{4pt}
\renewcommand{\arraystretch}{1.16}
\begin{longtable}{@{}>{\raggedright\arraybackslash}p{\dimexpr0.2\linewidth-2\tabcolsep\relax}>{\raggedright\arraybackslash}p{\dimexpr0.8\linewidth-2\tabcolsep\relax}@{}}
\caption{Inference configurations used with the official implementations. R, L, and K denote RoboTwin, LIBERO, and G1. Each row groups the settings for one evaluated system; platform-specific values are marked explicitly. History/conditioning frames, future frames, and sampling iterations have distinct meanings. Guidance values follow the corresponding sampler's convention.}\label{tab:wfb:inference}\\
\toprule
\textbf{Model} & \textbf{Applied inference configuration} \\
\midrule
\endfirsthead
\multicolumn{2}{l}{\small\textit{Table~\thetable{} (continued)}}\\
\toprule
\textbf{Model} & \textbf{Applied inference configuration} \\
\midrule
\endhead
\midrule
\multicolumn{2}{r}{\small\textit{Continued on next page}}\\
\endfoot
\bottomrule
\endlastfoot
BWM (R) & \textbf{Image input:} Head-camera view. \textbf{Sampling:} 50 steps; classifier-free guidance (CFG) scale 1. \textbf{Temporal and action interface:} 21-frame window including 9 history frames. Forward kinematics converts joint targets into bimanual end-effector poses and gripper commands. \\
Ctrl-World & \textbf{Image input:} 192 \(\times\) 320. \textbf{Sampling:} 25 steps; guidance 1. \textbf{Temporal and action interface:} 6 history frames and 5 output frames; 5 Hz. Actions use training-set first- and ninety-ninth-percentile normalization statistics. \\
Cosmos-Predict-2.5 & \textbf{Image input:} L: 256 \(\times\) 256; R/K: 256 \(\times\) 320. \textbf{Sampling:} R: 35 steps; L/K: 25 steps; guidance 0. \textbf{Temporal and action interface:} 1 conditioning frame and 12 future frames; 5 Hz. Platform-specific 7D/14D/26D action conditioning. The consumed prediction prefix follows the actual input action length. \\
DreamDojo & \textbf{Image input:} 480 \(\times\) 640. \textbf{Sampling:} 25 steps; guidance 0. \textbf{Temporal and action interface:} 1 conditioning frame and 12 future frames; 5 Hz. Actions map to a 384D conditioning interface. In R, each block of four actions subtracts its corresponding anchor. \\
DW05 & \textbf{Image input:} L: letterboxed to 384 \(\times\) 320. \textbf{Sampling:} R: 20 steps; L/K: 25 steps. \textbf{Temporal and action interface:} 1 conditioning frame and 8 future frames. L: 5 Hz; R/K: 7.5 Hz. L/K actions are padded to 32D. \\
A2World & \textbf{Image input:} 256 \(\times\) 256. \textbf{Sampling:} 20 steps; guidance 0. \textbf{Temporal and action interface:} 1 conditioning frame and 20 future frames. L: 20 Hz; R/K: 5 Hz. The L adapter maps 7D policy actions to the original simulation model's 14D conditioning interface. \\
GE-Sim & \textbf{Image input:} 256 \(\times\) 256. \textbf{Sampling:} 50 Euler steps. \textbf{Temporal and action interface:} 1 conditioning frame and 20 future frames, with 4 additional memory images. L: 20 Hz; R/K: 5 Hz. Images and robot states are jointly predicted. \\
\end{longtable}
\endgroup

Action-dimensionality changes occur within the world-model adapters; policies retain their native platform-specific action spaces.

\subsubsection{Fine-tuning and official implementations}
\label{app:wfb:training}

For each evaluated platform, we initialize each world model from an official checkpoint and fine-tune it for 60,000 steps on shared expert data, using the same training/test split across models. Training methods, trainable components, and optimization settings follow each model's official implementation, with platform-specific observation and action adapters. All reported results use the resulting platform-adapted checkpoints. Table~\ref{tab:wfb:resources} lists the official implementations and initialization sources.

Each world-model fine-tuning run and each closed-loop rollout worker uses eight NVIDIA A800 GPUs with 80\,GB memory each. The schedule standardizes update count and hardware allocation; batch size and computation per update remain model-specific, so training FLOPs and data exposure need not be equal.

\begingroup
\small
\setlength{\tabcolsep}{4pt}
\renewcommand{\arraystretch}{1.16}
\begin{longtable}{@{}>{\raggedright\arraybackslash}p{\dimexpr0.18\linewidth-2\tabcolsep\relax}>{\raggedright\arraybackslash}p{\dimexpr0.25\linewidth-2\tabcolsep\relax}>{\raggedright\arraybackslash}p{\dimexpr0.57\linewidth-2\tabcolsep\relax}@{}}
\caption{Official implementations and released checkpoint sources for platform-specific fine-tuning.}\label{tab:wfb:resources}\label{tab:official-model-resources}\\
\toprule
\textbf{Model} & \textbf{Official implementation} & \textbf{Initialization source} \\
\midrule
\endfirsthead
\multicolumn{3}{l}{\small\textit{Table~\thetable{} (continued)}}\\
\toprule
\textbf{Model} & \textbf{Official implementation} & \textbf{Initialization source} \\
\midrule
\endhead
\midrule
\multicolumn{3}{r}{\small\textit{Continued on next page}}\\
\endfoot
\bottomrule
\endlastfoot
BWM & \href{https://github.com/boundless-large-model/boundless-world-model}{Implementation} & \href{https://huggingface.co/BLM-Lab/Boundless-World-Model}{BWM checkpoint} \\
Ctrl-World & \href{https://github.com/Robert-gyj/Ctrl-World}{Implementation} & \href{https://huggingface.co/yjguo/Ctrl-World}{DROID-trained checkpoint} \\
Cosmos-Predict-2.5 & \href{https://github.com/nvidia-cosmos/cosmos-predict2.5}{Implementation} & \href{https://huggingface.co/nvidia/Cosmos-Predict2.5-2B/tree/main/robot/action-cond}{2B action-conditioned checkpoint} \\
DreamDojo & \href{https://github.com/NVIDIA/DreamDojo}{Implementation} & \href{https://huggingface.co/nvidia/DreamDojo/tree/main/2B_pretrain}{2B pretrained checkpoint} \\
DW05 & \href{https://github.com/dexmal/opendw}{OpenDW implementation} & \href{https://huggingface.co/Dexmal/DW05-Base}{Base checkpoint}; \href{https://huggingface.co/Dexmal/DW05-Robotwin}{RoboTwin-adapted checkpoint} \\
A2World & \href{https://github.com/LogosRoboticsGroup/A2World}{Implementation} & \href{https://huggingface.co/Fleurrr/A2World-World-Model}{Pretrained and LIBERO-adapted checkpoints} \\
GE-Sim & \href{https://github.com/AgibotTech/GE-Sim-V2}{V2 implementation} & \href{https://huggingface.co/agibot-world/Genie-Envisioner-Sim-v2.0/tree/main/checkpoints/gesim_community_v2.0.1_g01op_distill_2B}{V2.0.1 distilled 2B checkpoint} \\
\end{longtable}
\endgroup

\subsection{Closed-Loop Rollout Setup}
\label{app:wfb:execution}
\label{app:contract}
\label{app:rollout_setup}

\subsubsection{Reference scenes and initial conditions}
\label{app:wfb:scenes}

\textbf{RoboTwin.} We use \texttt{demo\_clean} with \texttt{seen} instructions. This setting retains task-specific randomization of object positions, orientations, and assets while disabling background, clutter, lighting, camera-position, and table-height perturbations. Candidate scene seeds start at 200000 and increase sequentially. Candidate scenes are checked using expert planning and task-success tests; accepted scenes are then reinitialized before executing the evaluated policy. We collect 50 accepted scenes for each task--policy pair.

The accepted seed sequences differ across RoboTwin policies. Episode indices are local to each task--policy collection. Within such a collection, all world models share the corresponding retained reference episodes. Cross-policy frequency comparisons aggregate each policy's task-level scene sample; they do not rely on episode-wise matching between policies.

\textbf{LIBERO.} For each task, we restore official initial-state indices 0--49 using collector seed 7 and perform ten settling steps before policy execution. These settling steps are excluded from the policy-action budget. We do not add pose or appearance perturbations to this initial-state set in the failure-type and failure-stage evaluations. Cross-policy scene pairing is established using the task definition, initial-state-file hash, and initial-state index.

\textbf{G1.} Each generated rollout is associated with a real execution through its task, policy version, and source UID. Initialization uses that execution's recorded initial RGB observation and 26D robot state. The source UID defines reference--model pairing. Different policy versions follow shared reset procedures but retain their own real-execution records.

\subsubsection{Reference--model pairing and random seeds}
\label{app:wfb:pairing}

Reference--model pairing fixes the task, policy checkpoint, source episode identity, initial RGB observation, initial robot state, and instruction. Within each task--policy setting, all models are scored on the same retained episode set.

Scene seeds identify candidate environments, policy seeds govern policy sampling, and generation seeds govern world-model sampling. The retained RoboTwin collection launchers specify default policy seeds of 0 for \(\pi_{0.5}\) and 42 for LingBot-VLA-v2 and FastWAM. Generation follows each implementation's seed schedule; BWM, for example, uses 42 + k for generation block k. The 50 reference episodes sample different accepted scenes or initial-state indices. These are scene-level samples rather than repeated trials of one fixed scene.

\subsubsection{Initialization, action consumption, and state feedback}
\label{app:wfb:feedback}

\textbf{Initialization.} Each world-model execution receives its source reference episode's initial RGB observation, robot state, and instruction. On RoboTwin, the BWM and DW05 implementations also use the first recorded reference action chunk for initialization. Reference-action conditioning is confined to initialization. Subsequent reference images, robot states, and actions do not condition generated execution; reference videos remain available for scoring.

\textbf{Closed-loop execution.} At each interaction round, the policy predicts an action sequence from the current observation and robot state, conditioned on the task instruction. Each world model advances the rollout according to its supported temporal configuration, then returns the corresponding observation and state for the next policy call. The final frame retained for the consumed action interval becomes the next observation, paired with the robot state at the same point on the action timeline.

\textbf{Robot-state feedback.} GE-Sim jointly predicts images and states; its predicted state corresponding to the retained image is fed to the policy. The other models propagate robot state from the consumed policy actions using their platform-specific update routines, including the corresponding coordinate and gripper conventions. These updates start from the recorded initial state and do not read subsequent reference states.

Reference execution advances through the simulator or physical robot using the same fixed policy checkpoint. After initialization, reference and generated executions obtain actions from their respective feedback streams. Their later action sequences may therefore differ even when their source initial conditions are matched.

\subsubsection{Rollout budget}
\label{app:wfb:horizons}

Following WorldArena~\citep{shang2026worldarena}, we set the world-model execution budget to 1.2 times the corresponding reference duration on all three platforms. This duration is converted to each model's action and frame timeline, with rounding to its supported temporal windows. To assess possible premature truncation on RoboTwin, we sample five failed episodes that reached the original rollout limit for each task--policy--world-model configuration (20 tasks, five policies, and seven world models), yielding 3,500 episodes. We extend the rollout horizon to the task-specific maximum episode length specified by the simulator. Only 17 episodes (0.49\%) succeed under the extended budget, indicating that failure outcomes in the sampled episodes are largely stable under longer evaluation horizons.

\subsection{Reference Repeatability (Ref--Ref)}
\label{app:reference_repeatability}
To assess the reproducibility of failure measurements, we independently rerun the policies in RoboTwin and LIBERO under the same evaluation settings. For failure-reason and failure-stage evaluation, we collect 50 new rollouts per task--policy setting from the same initial scenes, matching the original reference collection. We compare the two reference collections using the same annotation procedure, metrics, and aggregation rules. The resulting Ref--Ref scores provide an empirical baseline for interpreting reference--model discrepancies relative to the variation observed between independent reference executions.

For failure-boundary evaluation, Ref--Ref compares success-rate
changes between two reference estimates at paired scene conditions
using OA and GE (Section~\ref{sec:task_boundary}).
The two estimates are obtained from independent batches of
reference executions collected after the boundary pairs and
their endpoint order have been fixed. Neither estimate uses
the discovery rollouts, and the pairs and endpoint order remain
unchanged for this comparison.
Tables~\ref{tab:boundary-response-fidelity}
and~\ref{tab:libero_boundary_results} report reference repeatability
separately for position, orientation, and size changes.

\clearpage

\section{Annotation Protocol}
\label{app:annotation}
This appendix defines failure labels and annotation decisions, followed by the outcome annotation prompt.

\subsection{Failure Categories and Annotation Rules}
\label{app:failure_categories}
The two interpretable failure types adapt the planning/execution distinction in REFLECT~\citep{liu2023reflect} and Guardian~\citep{pacaud2025guardian} to trajectory annotation. Labels describe observable behavior in the context of the task instruction, without establishing internal policy causes. The examples below illustrate the definitions and are not additional measured trials.

\paragraph{Task Planning Failure.}
This category concerns task-inappropriate choices of targets, subgoals, or their ordering. It includes omitting a required step or attempting a subgoal before its prerequisites are satisfied. For example, closing a cabinet before inserting the required object, then attempting insertion through the closed door, violates a task dependency even if the closing motion succeeds.

\paragraph{Action Execution Failure.}
This category concerns failure to carry out an intended subgoal through motion or physical interaction. Examples include missing a grasp, dropping a held object during transport, and inaccurate placement. For example, the gripper approaches the intended mug but closes beside its handle and lifts without the mug. The observed interaction fails to achieve the intended grasp.

\paragraph{Indeterminate Failure.}
A confirmed failure is indeterminate when its primary type cannot be identified reliably, for example because the decisive interaction is occluded. This label is recorded as \texttt{Unjudgable}. These episodes remain in the all-episode denominator. TFE and TPC evaluate Task Planning and Action Execution.

\begin{figure}[!t]
\centering
\includegraphics[width=\linewidth,height=0.78\textheight,keepaspectratio]{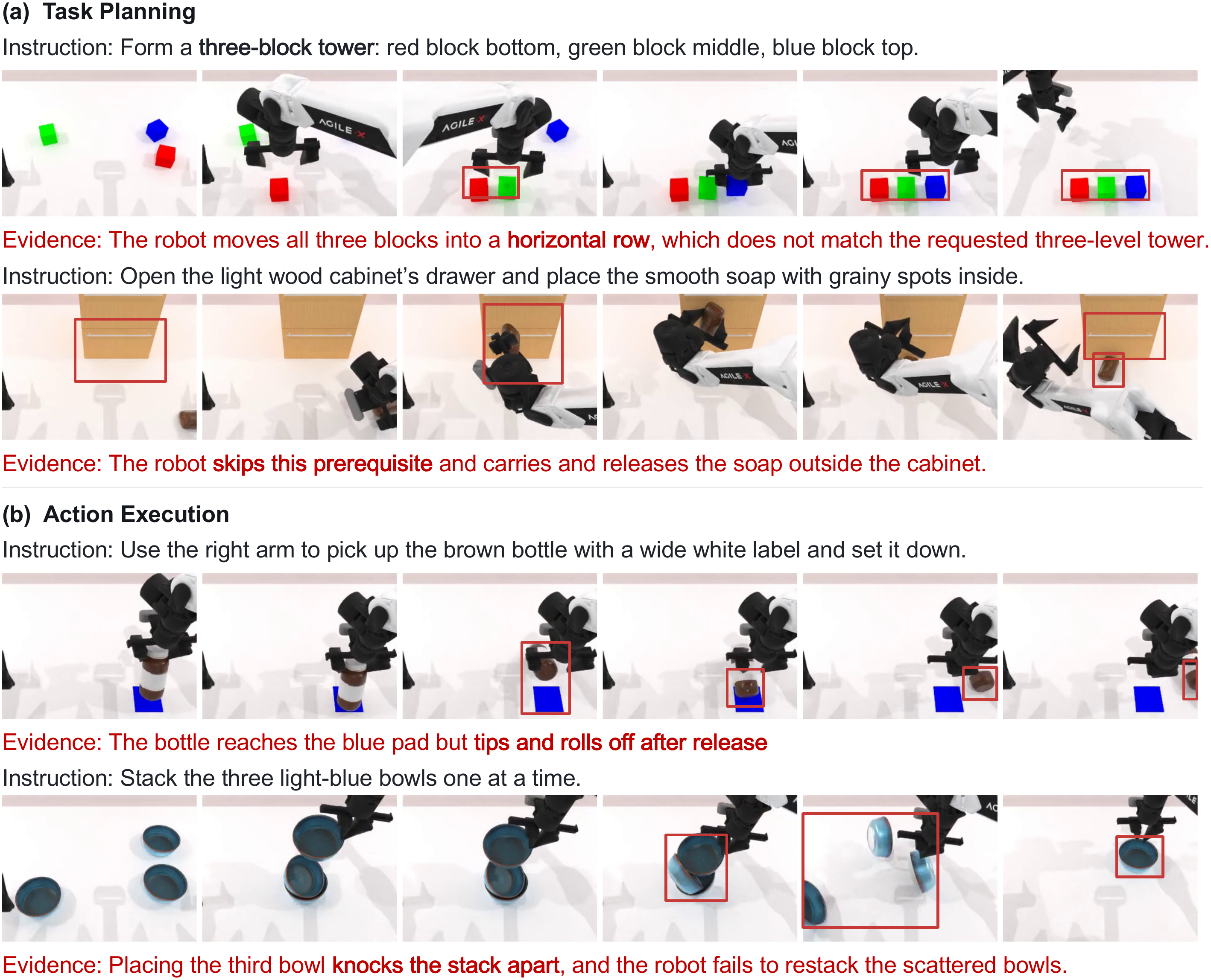}
\caption{Examples of task-planning and action-execution failures. (a) The robot arranges blocks in a row instead of the instructed tower, or attempts to place soap without first opening the drawer. (b) A released bottle tips and rolls away, or placing the third bowl disrupts the stack. Frames proceed from left to right; red boxes mark the relevant objects and interactions. These cases illustrate the two interpretable failure categories.}
\label{fig:failure_type_examples}
\end{figure}

Figure~\ref{fig:failure_type_examples} illustrates observable task-planning and action-execution failures with task instructions and frame sequences.

\subsection{Annotation Decisions and Observation Timing}
\label{app:annotation_decisions}
\paragraph{Outcome assessment and verification.}
Final outcome labels for reference executions in RoboTwin and LIBERO follow the simulator's native task predicates. World-model rollouts and real-world G1 executions are assessed from video by two VLMs, with one expert adjudicating disagreements. To verify consistency between video-based and simulator-based outcome assessment, we compare video annotations with native task predicates on a random 10\% sample of reference simulation videos, obtaining 99.8\% agreement.

\paragraph{Type and stage assignment.}
For each confirmed failure, two experts label one primary type and one decisive stage; a third expert adjudicates disagreements. When several errors occur, the type follows the earliest clearly supported primary cause; the stage identifies the earliest decisive event preventing completion. Stage names are copied from the task-specific list. Timestamps and critical-moment descriptions identify repeated occurrences; stage categories follow the task-specific decomposition in Appendix~\ref{app:stage_decomposition}. If a stage cannot be identified, annotators record it as \texttt{unknown}. Reference and model executions use the same rules; successful episodes contribute zero failure counts.

\clearpage
\subsection{Outcome Annotation Prompt}
\label{app:annotation_prompts}
\label{app:prompt_outcome}
The following prompt specifies video-based outcome evaluation. The placeholders \texttt{\{task\}} and \texttt{\{instruction\}} supply the task name and language instruction. The prompt requests a JSON response.

\begingroup
\definecolor{promptink}{HTML}{283E51}
\definecolor{promptborder}{HTML}{CDD5DD}
\definecolor{promptshade}{HTML}{F3F5F7}
\tcbset{
  annotationprompt/.style={enhanced,breakable,colback=white,colframe=promptborder,
    boxrule=0.45pt,arc=1mm,boxsep=0pt,left=10pt,right=10pt,top=8pt,bottom=8pt,
    before skip=5pt,after skip=8pt,fontupper=\fontsize{9}{11.2}\selectfont,
    fonttitle=\sffamily\bfseries\small,coltitle=promptink,colbacktitle=promptshade,
    toptitle=4pt,bottomtitle=4pt,
    title after break={\promptcontinuation\ (continued)}},
  promptinputs/.style={enhanced,colback=promptshade,colframe=promptshade,
    boxrule=0pt,arc=0.6mm,boxsep=0pt,left=7pt,right=7pt,top=5pt,bottom=5pt,
    before skip=2pt,after skip=3pt},
  promptjson/.style={enhanced,colback=promptshade,colframe=promptshade,
    boxrule=0pt,arc=0.6mm,boxsep=0pt,left=7pt,right=7pt,top=5pt,bottom=5pt,
    before skip=4pt,after skip=0pt}
}
\newcommand{\promptheading}[1]{\par\addvspace{7pt}\noindent{\color{promptink}\sffamily\bfseries\fontsize{8.5}{10}\selectfont\MakeUppercase{#1}}\par\nobreak\vspace{3pt}}
\newcommand{\promptcategory}[1]{\par\addvspace{6pt}\noindent{\color{promptink}\bfseries #1}\par\nobreak\vspace{2pt}}
\setlength{\parindent}{0pt}
\setlength{\parskip}{3pt}

\def\promptcontinuation{Outcome evaluation}
\begin{tcolorbox}[annotationprompt]
\promptheading{Role and objective}
You are a robotic task video evaluator. Determine whether the task was successfully completed based on the task instruction and the visible outcomes in the video.\par
\promptheading{Inputs}
\begin{tcolorbox}[promptinputs]
\textbf{Task name:} \texttt{\{task\}}\par
\textbf{Task instruction:} \texttt{\{instruction\}}\par
\end{tcolorbox}
\promptheading{Evaluation rules}
\begin{enumerate}[label=\textbf{\arabic*.},leftmargin=16pt,labelsep=5pt,itemsep=3pt,topsep=2pt,parsep=0pt]
\item Determine success by whether a visually identifiable task-completion state is achieved during the execution. The initial state alone does not establish success for a new execution. Once completion is observed, set \texttt{success=true}, even if the state changes afterward or the video continues. Do not assess whether the execution procedure was proper.
\item A blurry but reasonably identifiable completion state is sufficient for \texttt{success=true}. If no completion state is observed and the visible evidence supports that the task remains incomplete at the end of the video, set \texttt{success=false}.
\item For generated videos, ignore visual artifacts when they do not prevent outcome assessment. Focus on whether the task goal was achieved. Lower confidence when uncertain.
\item Rely only on the task instruction and visible evidence in the video. Do not infer the outcome from success/failure folder names or model names.
\item If the available visual evidence is insufficient to determine whether the task was completed, set \texttt{success="unjudgeable"}. Blur, a limited camera view, or an object being out of view does not by itself imply failure; use "unjudgeable" when these limitations prevent a reliable outcome judgment.
\item \texttt{success\_timestamp\_sec} is the approximate time in seconds when the task completion state first appears. Set it to null when \texttt{success=false} or \texttt{success=unjudgeable}.
\item Include only visually observable evidence in evidence. Do not classify the failure type or failure stage in this evaluation.
\end{enumerate}
In the JSON response, success must be true, false, or the string "unjudgeable".\par
\promptheading{Output format}
Return only valid JSON, without Markdown:
\begin{tcolorbox}[promptjson]
\begin{Verbatim}[fontsize=\footnotesize,breaklines=true,breakanywhere=true,breaksymbolleft={},breaksymbolright={},baselinestretch=1]
{
  "success": true,
  "success_timestamp_sec": 0.0,
  "reason": "<rationale and uncertainty>",
  "evidence": [
    "<visible evidence>"
  ],
  "confidence": 0.0
}
\end{Verbatim}
\end{tcolorbox}

\end{tcolorbox}

\endgroup

\clearpage
\section{Boundary Construction and Interventions}
\label{app:boundary}
We construct 60 boundary pairs in each of RoboTwin and LIBERO, with 20 per factor. Starting from successful scenes, we intervene on one object's position, orientation, or size while holding the instruction, policy weights, and remaining scene settings fixed. RoboTwin pairs use LingBot-VLA-v2, $\pi_{0.5}$, and FastWAM; LIBERO uses the full-data, data-50, and data-20 $\pi_{0.5}$ variants.

\subsection{Intervention Ranges}
Table~\ref{tab:app_boundary_ranges} distinguishes RoboTwin's preset search ranges from its selected endpoints. Position and rotation are offsets from the initial scene; size multipliers are relative to the object's original linear dimensions. On RoboTwin, cube rotation paths are restricted to $\pm40^{\circ}$, and the middle object in size-ranking tasks permits only multipliers 0.9, 1.0, and 1.1; these special paths contribute no final pairs. For endpoint multipliers $s_1$ and $s_2$, the relative size change is $100|s_2/s_1-1|\%$.

LIBERO uses the same preset search ranges, with 2\,cm position steps and 0.1 scale-factor steps. Its yaw step is $10^{\circ}$, compared with RoboTwin's $20^{\circ}$. Both platforms follow the discovery and evaluation procedure below.

\begin{table}[!htbp]
\centering\small
\setlength{\tabcolsep}{3pt}
\renewcommand{\arraystretch}{1.13}
\caption{RoboTwin boundary search ranges, grid steps, and selected endpoint ranges. Position and yaw are offsets from the initial scene; size is a linear scale multiplier. LIBERO shares the search ranges and uses 2\,cm position steps, a $10^{\circ}$ yaw step, and 0.1 scale-factor steps.}
\label{tab:app_boundary_ranges}
\begin{tabularx}{\linewidth}{@{}>{\raggedright\arraybackslash}Xccc@{}}
\toprule
Intervention & \shortstack{Search\\range} & \shortstack{Pair\\step} & \shortstack{Selected endpoint\\range}\\
\midrule
Planar position ($x$ or $y$) & $-40$ to $40$\,cm & 4\,cm & $-36$ to $28$\,cm\\
Yaw about the vertical axis & $-120^{\circ}$ to $120^{\circ}$ & $20^{\circ}$ & $-120^{\circ}$ to $120^{\circ}$\\
Linear size multiplier & 0.4--2.0 & 0.2 & 0.4--1.8\\
\bottomrule
\end{tabularx}
\end{table}

\subsection{Boundary Discovery and Independent Evaluation}
Both platforms separate boundary discovery from final reference estimation using the following procedure.

\paragraph{(1) Discover candidate boundaries.}
We estimate success at adjacent conditions on each preset path using ten complete reference rollouts per endpoint. A pair qualifies when the success rates differ by at least 30 percentage points, equivalent to a difference of three successes out of ten. Both increases and decreases along the scan direction qualify. Physical pairs are deduplicated, and 20 pairs are selected for each intervention factor in each platform.

\paragraph{(2) Freeze the evaluation set.}
Before independent reference re-estimation, we fix all 60 pairs per platform, including their tasks, policies, base scenes, and endpoint conditions. Endpoints are ordered from higher to lower success according to the discovery estimates. The set and endpoint order remain fixed throughout evaluation: no pair is removed or reordered based on the re-estimated success rates, even if its difference falls below 30 percentage points or reverses direction.

\paragraph{(3) Independently re-estimate reference success.}
At every frozen endpoint in both platforms, we collect ten new reference executions using random seeds disjoint from those used for discovery. Success follows the simulator's task predicate. All complete evaluation executions enter the estimate; exploratory probes, invalid scenes, and infrastructure errors are not counted as task failures. Only these ten new executions determine the reference success rate used for final evaluation; discovery executions are not pooled into this estimate.

\paragraph{(4) Evaluate world models.}
Each world model runs the same policy ten times at each frozen endpoint, from the corresponding initialization. OA and GE (Section~\ref{sec:task_boundary}) compare model success-rate changes with the independently re-estimated reference changes, using the frozen endpoint order. The reported fraction of underestimated reference changes also uses these independent estimates.

\subsection{Scene Examples}
\begin{figure}[!htbp]
\centering
\includegraphics[width=\linewidth,height=0.78\textheight,keepaspectratio]{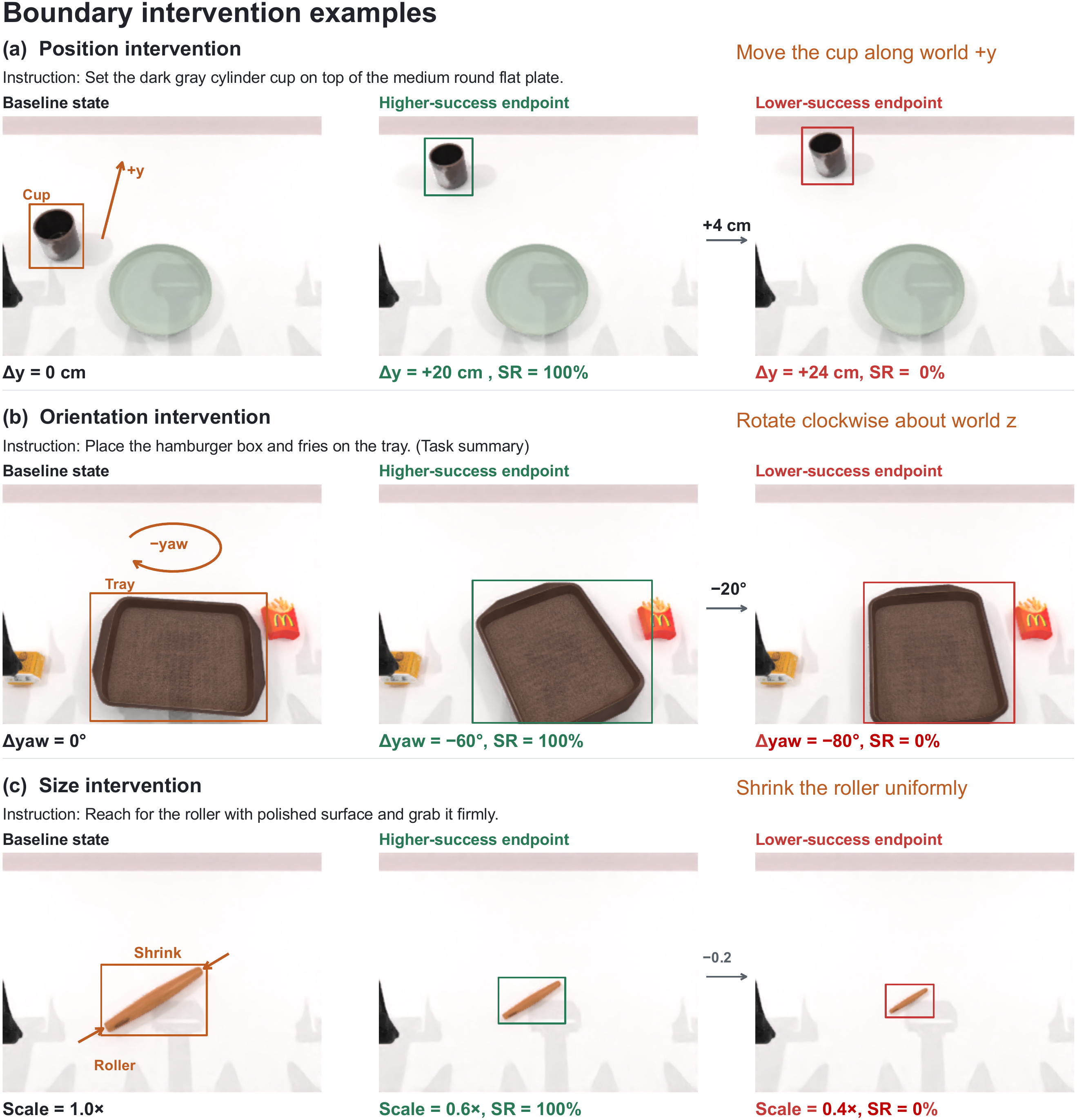}
\caption{Scene interventions for boundary construction. SR denotes success rate; $\Delta y$ and $\Delta\mathrm{yaw}$ are offsets from the base scene, with scale factors relative to the original object size. The conditions differ by (a) cup translation along world $+y$, from $+20$ to $+24$\,cm (success: 100\% to 0\%); (b) tray rotation about world $z$, from $-60^{\circ}$ to $-80^{\circ}$ (100\% to 0\%); or (c) uniform roller scaling, from $0.6\times$ to $0.4\times$ (100\% to 0\%).}
\label{fig:boundary_intervention_examples}
\end{figure}

Figure~\ref{fig:boundary_intervention_examples} illustrates RoboTwin's position, orientation, and size interventions. Each row changes one scene factor while keeping the task and policy fixed.

\FloatBarrier

\clearpage
\section{Additional Experimental Results}
\label{app:additional_results}

\subsection{RoboTwin Failure-Type Policy Correlation}
\label{app:robotwin_type_ranking}
Table~\ref{tab:robotwin_type_rank} reports planning, execution, and aggregate TPC on RoboTwin.
\begin{table}[!htbp]
\centering\small
\setlength{\tabcolsep}{5pt}
\renewcommand{\arraystretch}{1.16}
\caption{Type-specific Policy Correlation (TPC) on RoboTwin ($\uparrow$). Type columns average defined within-task Pearson correlations across five policies. Overall Mean averages defined type correlations within tasks, then tasks equally. Bold marks the best world-model values.}
\label{tab:robotwin_type_rank}
\begin{tabular}{@{}lrrr@{}}
\toprule
Evaluator & Task Planning & Action Execution & Overall Mean\\
\midrule
Ref--Ref & $0.796$ & $0.831$ & $0.814$\\
\midrule
A2World & $0.353$ & $-0.086$ & $0.134$\\
BWM & $-0.121$ & $0.143$ & $0.011$\\
Cosmos & $0.008$ & $0.038$ & $0.023$\\
Ctrl & $0.177$ & $0.204$ & $0.191$\\
DW05 & $\mathbf{0.456}$ & $\mathbf{0.348}$ & $\mathbf{0.402}$\\
DreamDojo & $0.102$ & $-0.306$ & $-0.102$\\
GE-Sim & $0.336$ & $0.142$ & $0.239$\\
\bottomrule
\end{tabular}
\end{table}

\FloatBarrier

\subsection{RoboTwin Failure-Stage Profiles Across Policies}
\label{app:robotwin_stage_profiles}
\begin{figure}[!htbp]
\centering
\includegraphics[width=\linewidth]{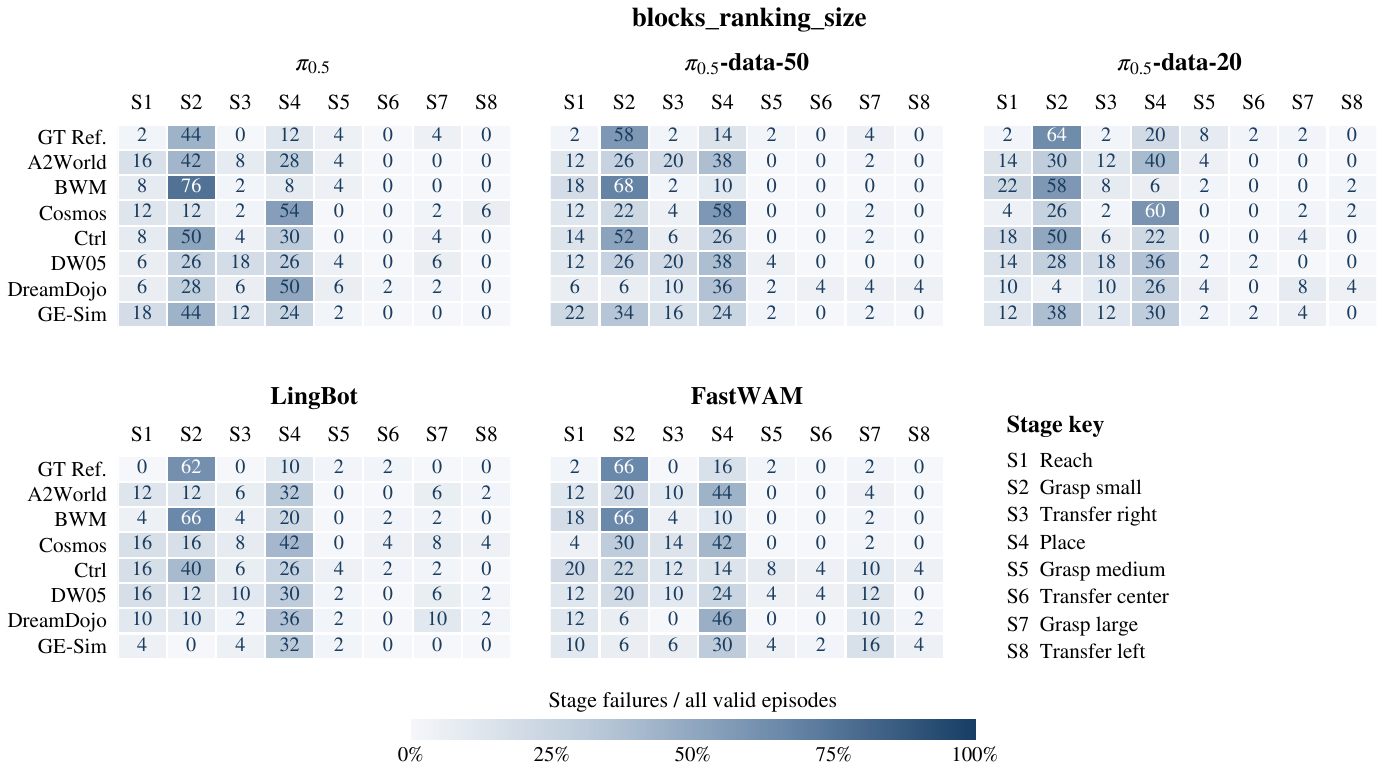}
\caption{Stage-specific failure frequencies for \texttt{blocks\_ranking\_size} across five policies. Each panel shows reference execution and seven world models; cells report the percentage of all valid episodes with a failure at each stage. The stage key and color scale are shared.}
\label{fig:robotwin_stage_heatmap_all_policies}
\end{figure}

Figure~\ref{fig:robotwin_stage_heatmap_all_policies} compares stage-specific failure frequencies across all five policies on \texttt{blocks\_ranking\_size}. For every policy, reference failures peak at Grasp small (S2), while DreamDojo's generated failures peak at Place (S4).
\FloatBarrier

\clearpage
\subsection{RoboTwin Failure-Boundary Responses}
\label{app:robotwin_boundary_responses}
\begin{figure}[!htbp]
\centering
\includegraphics[width=\linewidth]{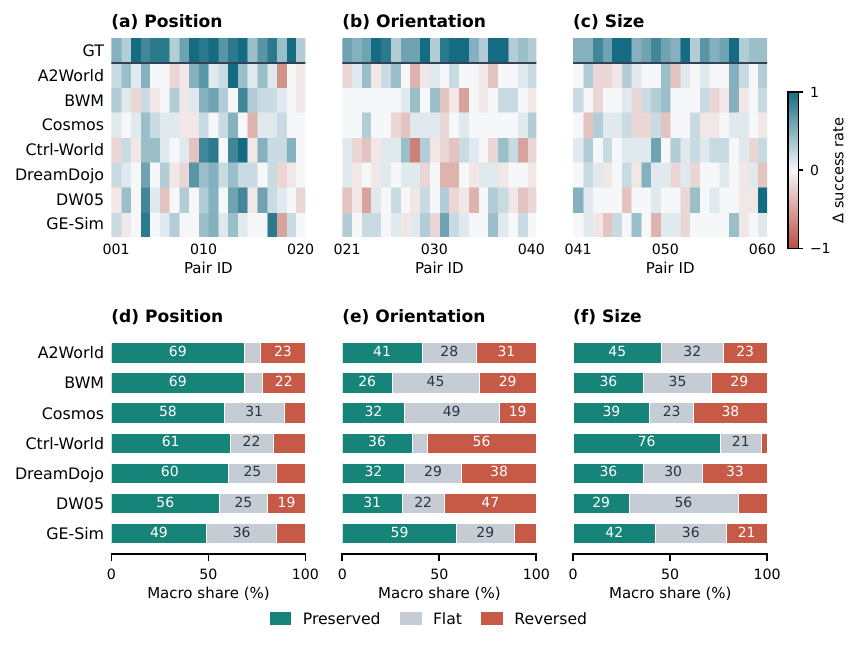}
\caption{Responses across 60 frozen boundary pairs. \textbf{(a--c)} Reference and model success-rate changes, labeled $\Delta^{\mathrm{GT}}$ and $\Delta^{\mathrm{WM}}$, respectively. Each column uses the endpoint order fixed during discovery; reference changes come from independent re-estimation. \textbf{(d--f)} Setting-weighted shares of preserved, flat, and reversed responses. Model changes within $\pm2$ percentage points are flat; nonflat changes are preserved or reversed according to agreement with the independently estimated reference direction. The 95.2\% underestimation rate is descriptive: the 420 model--pair comparisons across seven models share boundary pairs and some endpoints and are not independent.}
\label{fig:boundary-response-diagnostics}
\end{figure}

Figure~\ref{fig:boundary-response-diagnostics} shows success-rate changes for individual boundary pairs and the shares of preserved, flat, and reversed responses for each intervention factor.
\FloatBarrier

\clearpage
\begingroup
\captionsetup{skip=3pt}
\setlength{\parskip}{2pt}
\setlength{\intextsep}{4pt}
\setlength{\textfloatsep}{4pt}
\setlength{\floatsep}{4pt}
\subsection{LIBERO Failure-Type Understanding}
\label{app:libero_type_results}
We evaluate six world models on LIBERO-10 using $\pi_{0.5}$ and its data-50 and data-20 variants. Table~\ref{tab:libero_tfe} reports TFE.

\begin{table}[!htbp]
\centering\small
\setlength{\tabcolsep}{5pt}
\renewcommand{\arraystretch}{1.03}
\caption{Type Frequency Error (TFE) on LIBERO ($\downarrow$). Policy columns average tasks; Mean weights the three policies equally. Bold marks the best world-model values.}
\label{tab:libero_tfe}
\begin{tabular}{@{}lrrrr@{}}
\toprule
Evaluator & $\pi_{0.5}$ & $\pi_{0.5}$-data-50 & $\pi_{0.5}$-data-20 & Mean\\
\midrule
Ref--Ref & $0.032$ & $0.056$ & $0.048$ & $0.045$\\
\midrule
A2World & $0.300$ & $0.264$ & $\mathbf{0.248}$ & $\mathbf{0.271}$\\
Cosmos & $0.322$ & $0.288$ & $0.254$ & $0.288$\\
Ctrl & $\mathbf{0.290}$ & $\mathbf{0.257}$ & $0.282$ & $0.276$\\
DW05 & $0.352$ & $0.293$ & $0.284$ & $0.310$\\
DreamDojo & $0.421$ & $0.384$ & $0.338$ & $0.381$\\
GE-Sim & $0.352$ & $0.324$ & $0.351$ & $0.342$\\
\bottomrule
\end{tabular}
\end{table}

\paragraph{Type Frequency Error (TFE).}
\textbf{Generated rollouts substantially misestimate the prevalence of policy failure types.} Mean TFE ranges from 0.271 to 0.381, compared with 0.045 for Ref--Ref. Ctrl has the lowest TFE for the full-data and data-50 policies, whereas A2World is most accurate for data-20: evaluator accuracy varies even within one policy family.

\begin{table}[!htbp]
\centering\small
\setlength{\tabcolsep}{5pt}
\renewcommand{\arraystretch}{1.03}
\caption{LIBERO TPC across three policies ($\uparrow$). Bold marks the best world-model values.}
\label{tab:libero_type_rank}
\begin{tabular}{@{}lr@{}}
\toprule
Evaluator & TPC\\
\midrule
Ref--Ref & $0.780$\\
\midrule
A2World & $-0.062$\\
Cosmos & $\mathbf{0.242}$\\
Ctrl & $-0.189$\\
DW05 & $0.043$\\
DreamDojo & $0.108$\\
GE-Sim & $0.029$\\
\bottomrule
\end{tabular}
\end{table}

\paragraph{Type-specific Policy Correlation (TPC).}
\textbf{World models poorly preserve cross-policy differences in failure-type frequencies.} TPC ranges from $-0.189$ to 0.242, below Ref--Ref's 0.780 (Table~\ref{tab:libero_type_rank}).
\FloatBarrier
\endgroup

\clearpage
\subsection{LIBERO Failure-Stage Understanding}
\label{app:libero_stage_results}
\begin{figure}[!htbp]
\centering
\includegraphics[width=\linewidth]{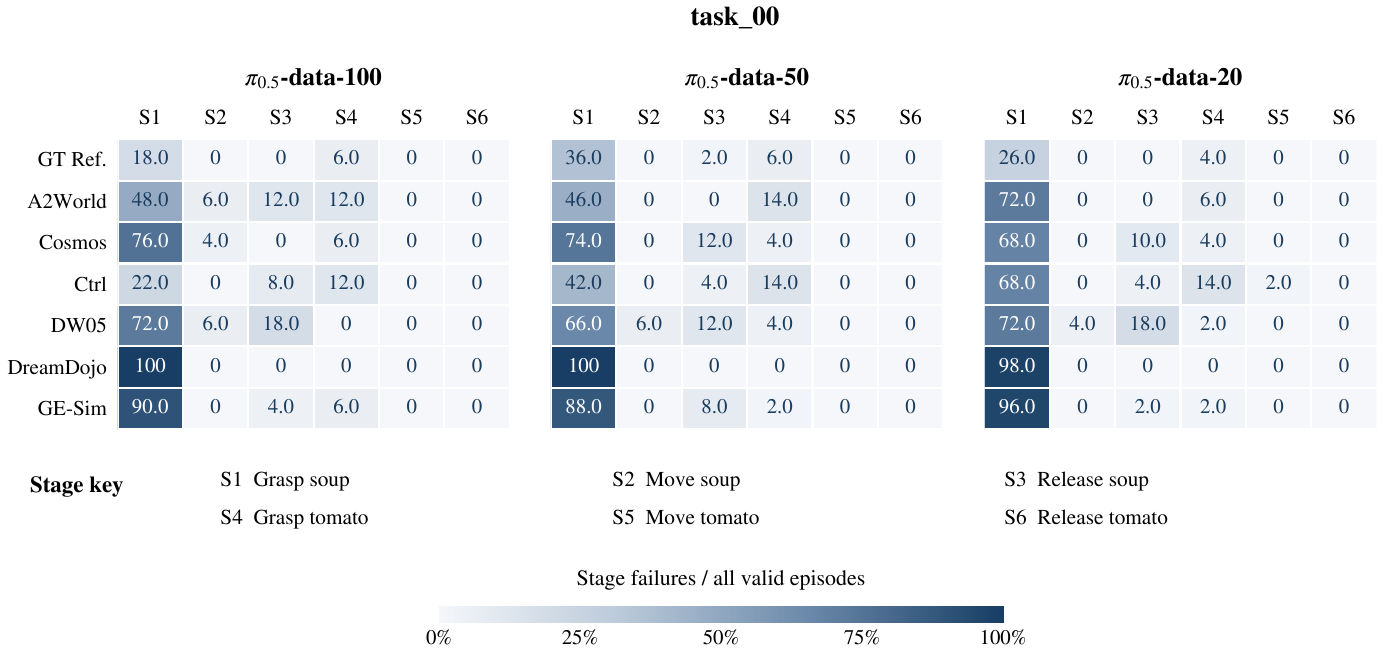}
\caption{Stage-specific failure frequencies for LIBERO \texttt{task\_00} across three policies. Each panel compares reference execution with six world models; cells give the percentage of all valid episodes with a failure at each stage. The stage key and color scale are shared. The data-100 policy is the full-data $\pi_{0.5}$ used in the text and tables.}
\label{fig:libero_stage_heatmap}
\end{figure}

Table~\ref{tab:libero_stage_results} reports SFE and SPC for the same three policy variants. Figure~\ref{fig:libero_stage_heatmap} illustrates stage-specific failure frequencies on \texttt{task\_00}, which requires placing both the alphabet soup and tomato sauce in a basket.

\begin{table}[!htbp]
\centering\small
\setlength{\tabcolsep}{5pt}
\renewcommand{\arraystretch}{1.16}
\caption{Stage Frequency Error (SFE, $\downarrow$) and Stage-specific Policy Correlation (SPC, $\uparrow$) on LIBERO. SFE policy columns average tasks; Mean averages policies within tasks, then tasks. Frequencies use all valid paired episodes. SPC averages defined within-task Pearson correlations over stages, then tasks. Bold marks the best world-model values.}
\label{tab:libero_stage_results}
\begin{tabular}{@{}lrrrrr@{}}
\toprule
& \multicolumn{4}{c}{SFE ($\downarrow$)} & SPC ($\uparrow$)\\
\cmidrule(lr){2-5}
Evaluator & $\pi_{0.5}$ & $\pi_{0.5}$-data-50 & $\pi_{0.5}$-data-20 & Mean & \\
\midrule
Ref--Ref & $0.064$ & $0.124$ & $0.140$ & $0.109$ & $0.680$\\
\midrule
A2World & $0.808$ & $0.796$ & $\mathbf{0.768}$ & $0.791$ & $-0.271$\\
Cosmos & $0.836$ & $0.832$ & $0.824$ & $0.831$ & $-0.173$\\
Ctrl & $\mathbf{0.692}$ & $\mathbf{0.682}$ & $0.844$ & $\mathbf{0.739}$ & $-0.021$\\
DW05 & $0.932$ & $0.862$ & $0.928$ & $0.907$ & $-0.235$\\
DreamDojo & $0.998$ & $0.988$ & $0.988$ & $0.991$ & $0.052$\\
GE-Sim & $0.856$ & $0.844$ & $0.938$ & $0.879$ & $\mathbf{0.228}$\\
\bottomrule
\end{tabular}
\end{table}

\paragraph{Stage Frequency Error and Stage-specific Policy Correlation.}
\textbf{Low frequency error does not ensure faithful cross-policy comparisons of stage-failure frequencies.} Ctrl has the lowest mean SFE (0.739), but its SPC is $-0.021$. Conversely, GE-Sim has the highest SPC (0.228) despite a larger SFE (0.879). Ref--Ref achieves SFE 0.109 and SPC 0.680, while four of six models have negative SPC. SFE and SPC capture complementary aspects of the full-task stage-failure profile: frequency accuracy and cross-policy variation.

\paragraph{Case study: exaggerated early-stage failures.}
\textbf{Correctly locating the dominant failure stage does not ensure accurate failure-frequency estimates.} In \texttt{task\_00}, reference failures at Grasp soup (S1) occur in 18\%, 36\%, and 26\% of episodes for the three policies, whereas DreamDojo predicts 100\%, 100\%, and 98\%. Although both identify S1 as the peak, the generated frequencies greatly overstate the prevalence of early grasp failures during full-task execution.
\FloatBarrier

\clearpage
\begingroup
\setlength{\parskip}{2pt}
\setlength{\intextsep}{8pt}
\setlength{\textfloatsep}{8pt}
\captionsetup{skip=4pt}
\subsection{LIBERO Failure-Boundary Understanding}
\label{app:libero_boundary_results}
Table~\ref{tab:libero_boundary_results} reports OA and GE for six world models and the Ref--Ref baseline on 60 LIBERO boundary pairs.

\begin{table}[!htbp]
\centering\small
\setlength{\tabcolsep}{5pt}
\renewcommand{\arraystretch}{1.13}
\caption{Failure-boundary results on LIBERO (OA $\uparrow$, GE $\downarrow$). Ref--Ref reports reference repeatability. Bold marks the best world-model value in each column, including ties.}
\label{tab:libero_boundary_results}
\begin{tabular}{@{}lrrrrrr@{}}
\toprule
Evaluator & \multicolumn{2}{c}{Position} & \multicolumn{2}{c}{Orientation} & \multicolumn{2}{c}{Size} \\
\cmidrule(lr){2-3}\cmidrule(lr){4-5}\cmidrule(lr){6-7}
 & OA $\uparrow$ & GE $\downarrow$ & OA $\uparrow$ & GE $\downarrow$ & OA $\uparrow$ & GE $\downarrow$ \\
\midrule
Ref--Ref & 0.950 & 0.035 & 0.850 & 0.080 & 0.900 & 0.065 \\
\midrule
A2World & 0.067 & \textbf{0.517} & 0.150 & 0.584 & 0.000 & 0.584 \\
Cosmos-Predict-2.5 & 0.067 & 0.522 & 0.075 & 0.534 & 0.000 & 0.556 \\
Ctrl-World & 0.000 & 0.528 & \textbf{0.175} & \textbf{0.514} & 0.077 & 0.543 \\
DreamDojo & 0.033 & 0.522 & 0.000 & 0.543 & 0.000 & 0.530 \\
DW05 & 0.000 & 0.528 & 0.000 & 0.555 & 0.000 & 0.549 \\
GE-Sim & \textbf{0.089} & \textbf{0.517} & 0.150 & 0.545 & \textbf{0.154} & \textbf{0.504} \\
\bottomrule
\end{tabular}
\end{table}

\paragraph{Failure-boundary evaluation.}
\textbf{(1) World models rarely identify the correct direction of policy sensitivity on LIBERO.} OA ranges from 0 to 0.175 across all model--factor combinations, limiting their ability to identify which nearby condition is less reliable. Ref--Ref instead attains OA of 0.85--0.95 across the three factors. \textbf{(2) World models do not reliably quantify how scene changes affect success rates.} Even the lowest world-model GE for each factor remains 0.504--0.517, corresponding to errors of 50.4--51.7 percentage points, compared with Ref--Ref GE of 0.035--0.080.

\begin{figure}[!htbp]
\centering
\includegraphics[width=0.55\linewidth]{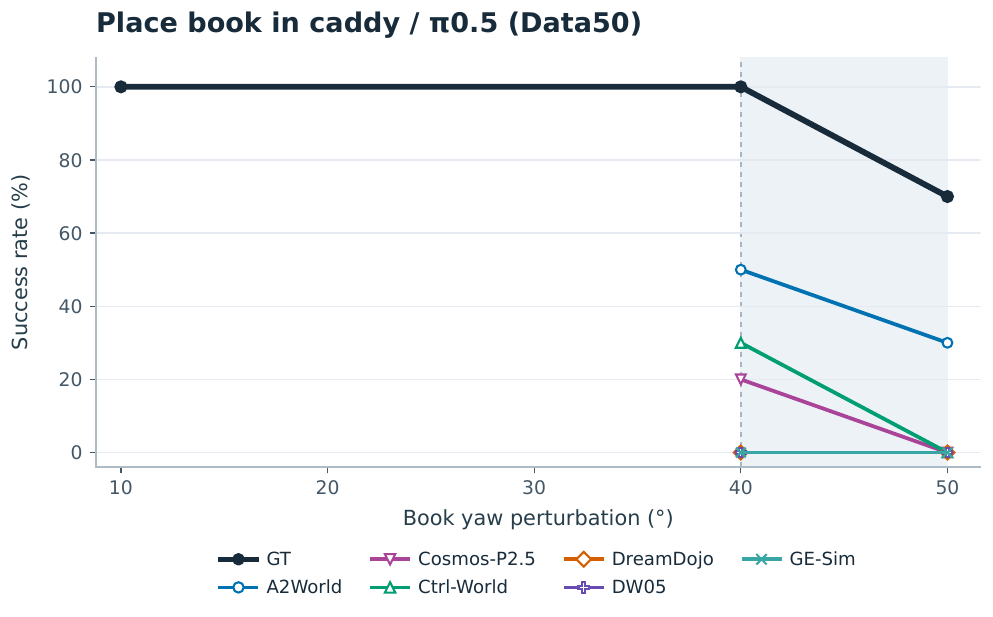}
\caption{Orientation case for $\pi_{0.5}$-data-50 on the LIBERO task of placing a book in a caddy. The reference curve shows sampled book yaw perturbations; world-model curves connect the highlighted $40^\circ$ and $50^\circ$ endpoints. Cosmos-P2.5 denotes Cosmos-Predict-2.5.}
\label{fig:libero_boundary_case}
\end{figure}

\paragraph{Case study: a missed orientation boundary.}
\looseness=-1
\textbf{Uniformly pessimistic predictions can hide differences between nearby conditions.} As book yaw increases from $40^\circ$ to $50^\circ$, reference success falls from 100\% to 70\%. DreamDojo, DW05, and GE-Sim instead predict 0\% at both endpoints (Figure~\ref{fig:libero_boundary_case}). These evaluators would label both conditions as uniformly unreliable, obscuring the reference policy's loss of reliability under the orientation change.
\FloatBarrier
\endgroup

\clearpage
\begingroup
\setlength{\parskip}{2pt}
\captionsetup{font=footnotesize,skip=3pt}
\setlength{\intextsep}{4pt}
\setlength{\textfloatsep}{4pt}
\setlength{\floatsep}{4pt}
\subsection{Real-World Results on G1}
\label{app:g1_failure_results}
\label{app:g1_outcomes}
\looseness=-1 We evaluate six world models on sweeping trash and collecting clothes, using four task-specific policy versions, V1--V4, for each task. Version labels do not imply a performance ordering. Table~\ref{tab:kairos-tfe-main} reports TFE and SFE. Table~\ref{tab:g1_reference_success_rates} gives reference success rates, with V3 performing best in both tasks.

\begin{table}[!htbp]
\centering\small
\caption{Failure-diagnostic errors for real-world tasks on G1 ($\downarrow$). TFE averages two failure-type errors; SFE sums four stage errors. Task scores average four policy versions; Mean weights the two tasks equally. Bold marks column minima.}
\label{tab:kairos-tfe-main}
\label{tab:kairos-sfe-main}
\setlength{\tabcolsep}{5pt}
\renewcommand{\arraystretch}{1.00}
\begin{tabular}{@{}lrrr@{\hspace{16pt}}rrr@{}}
\toprule
& \multicolumn{3}{c}{\textbf{Failure type: TFE}} & \multicolumn{3}{c}{\textbf{Failure stage: SFE}}\\
\cmidrule(lr){2-4}\cmidrule(l){5-7}
World model & Sweep & Clothes & Mean & Sweep & Clothes & Mean\\
\midrule
Ctrl & 0.0971 & 0.1195 & 0.1083 & 0.8284 & 0.4042 & 0.6163\\
DreamDojo & 0.1528 & 0.1715 & 0.1621 & 0.8340 & 0.5800 & 0.7070\\
Cosmos & 0.1796 & 0.2929 & 0.2362 & 0.7143 & 0.9544 & 0.8343\\
DW05 & 0.1429 & 0.2171 & 0.1800 & 0.6436 & 0.7269 & 0.6852\\
A2World & \textbf{0.0603} & 0.1258 & 0.0930 & \textbf{0.2759} & 0.3975 & \textbf{0.3367}\\
GE-Sim & 0.0838 & \textbf{0.0992} & \textbf{0.0915} & 0.5620 & \textbf{0.3829} & 0.4725\\
\bottomrule
\end{tabular}
\end{table}

\paragraph{Type and Stage Frequency Errors.}
\textbf{The most accurate evaluator depends on the task and the evaluation dimension.} A2World has the lowest sweeping TFE and SFE (0.0603 and 0.2759), while GE-Sim has the lowest clothes-collection errors (0.0992 and 0.3829). Across tasks, GE-Sim has the lowest mean TFE (0.0915), whereas A2World has the lowest mean SFE (0.3367).

\begin{table}[!htbp]
\centering
\small
\caption{Task-specific Pearson agreement across four policy versions for real-world tasks on G1. TPC and SPC average defined type and stage correlations, respectively. Bold marks column maxima.}
\label{tab:kairos-ranking}
\setlength{\tabcolsep}{7pt}
\renewcommand{\arraystretch}{1.03}
\begin{tabular}{lrrrr}
\toprule
 & \multicolumn{2}{c}{Sweeping trash} & \multicolumn{2}{c}{Collecting clothes} \\
\cmidrule(lr){2-3}\cmidrule(lr){4-5}
Model & TPC ($\uparrow$) & SPC ($\uparrow$) & TPC ($\uparrow$) & SPC ($\uparrow$) \\
\midrule
Ctrl & 0.5016 & -0.1270 & \textbf{0.6812} & 0.3463 \\
DreamDojo & -0.4996 & 0.1648 & 0.3714 & \textbf{0.4824} \\
Cosmos & -0.6619 & -0.2548 & 0.1005 & 0.2258 \\
DW05 & 0.1046 & -0.0251 & 0.0569 & 0.1479 \\
A2World & \textbf{0.5118} & \textbf{0.4667} & -0.2089 & 0.0545 \\
GE-Sim & 0.4406 & -0.2479 & 0.3014 & 0.2336 \\
\bottomrule
\end{tabular}
\end{table}

\begin{table}[!htbp]
\centering\small
\setlength{\tabcolsep}{5pt}
\renewcommand{\arraystretch}{1.00}
\caption{Reference success rates (\%) for real-world tasks on G1, computed over executions with judged outcomes. Each task has four policy versions; bold marks the highest rate within each task.}
\label{tab:g1_reference_success_rates}
\begin{tabular}{lrr}
\toprule
Policy & Sweeping trash & Collecting clothes\\
\midrule
V1 & 30.61 & 28.00\\
V2 & 50.00 & 59.18\\
V3 & \textbf{65.31} & \textbf{78.00}\\
V4 & 28.00 & 56.25\\
\bottomrule
\end{tabular}

\end{table}

\paragraph{Type- and Stage-specific Policy Correlation.}
\looseness=-1
\textbf{(1) Evaluator fidelity does not transfer uniformly between real-world tasks.} A2World has the highest sweeping TPC and SPC (0.5118 and 0.4667), but drops to $-0.2089$ and 0.0545 for clothes collection (Table~\ref{tab:kairos-ranking}). \textbf{(2) Preserving type differences does not ensure faithful comparisons of stage-failure frequencies.} GE-Sim has a sweeping TPC of 0.4406 but an SPC of $-0.2479$. For clothes collection, Ctrl has the highest TPC (0.6812), whereas DreamDojo has the highest SPC (0.4824). Evaluator suitability therefore depends on both the task and the diagnostic dimension.
\FloatBarrier
\endgroup

\clearpage
\section{Qualitative Reference and World-Model Comparisons}
\label{app:qualitative_comparisons}
The following examples illustrate preserved failures, failure-type mismatches, and failure-stage mismatches through reference and generated frame sequences. The displayed annotations identify the behaviors and stages used in each comparison.

\subsection{Simulation}
\begin{figure}[!htbp]
\centering
\includegraphics[width=\linewidth,height=0.64\textheight,keepaspectratio]{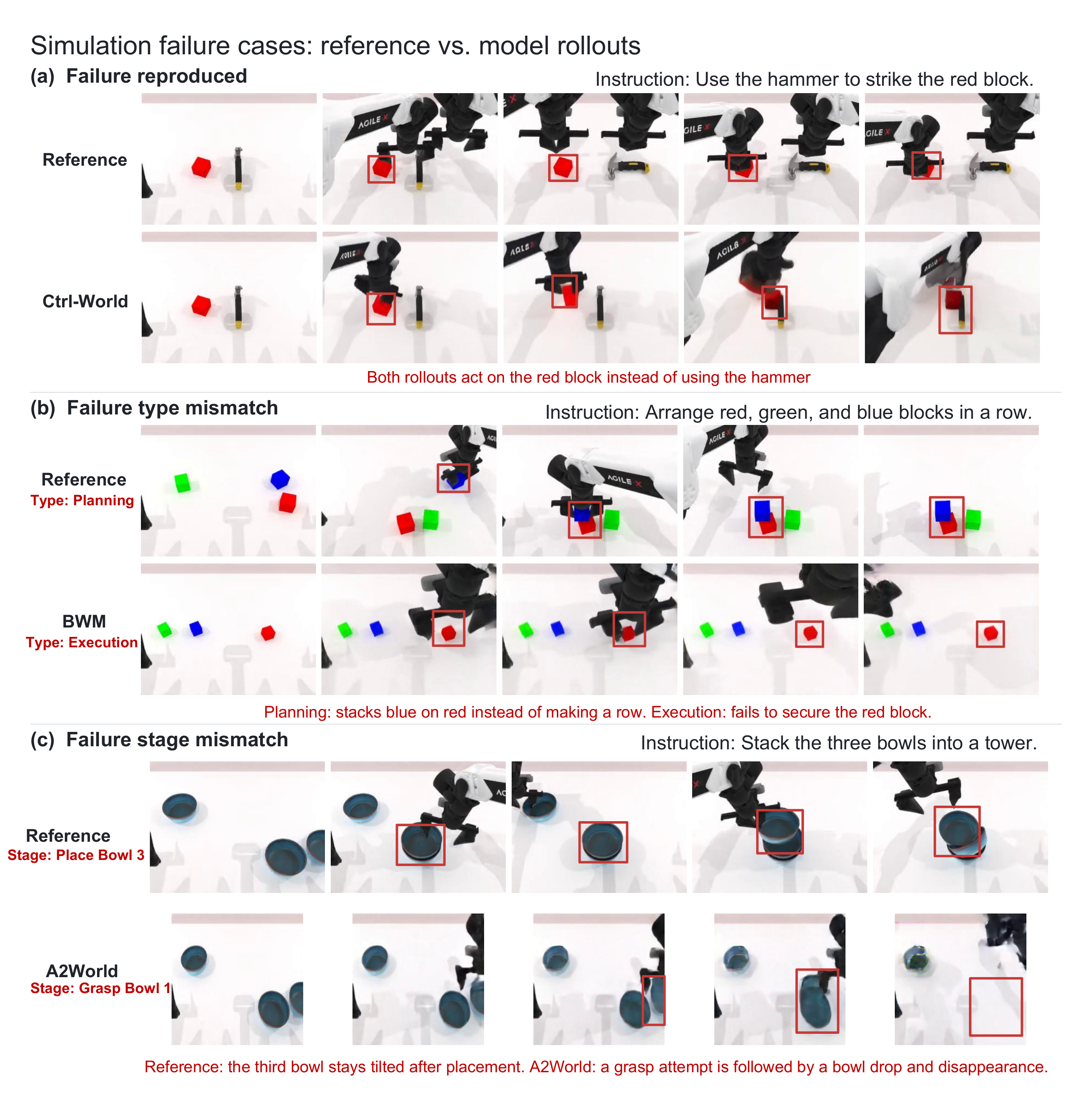}
\caption{Qualitative failure comparisons between reference and world-model rollouts in simulation. (a) Reference and Ctrl-World both manipulate the red block directly instead of using the hammer. (b) The blocks-ranking-color example shows different annotated failure types: the reference arranges the blue block on top of the red one in instead of in a row, while the model fails to acquire the red block. (c) The reference fails during placement of the third bowl, while A2World fails during acquisition of the first. Frames proceed from left to right, with red boxes marking the displayed evidence.}
\label{fig:simulation_failure_comparisons}
\end{figure}

Figure~\ref{fig:simulation_failure_comparisons} compares reference executions with Ctrl-World, BWM and A2World on tool use, blocks arrangement, and bowl stacking.

\FloatBarrier

\subsection{Real-World Tasks on G1}
\begin{figure}[!htbp]
\centering
\includegraphics[width=0.80\linewidth,height=0.68\textheight,keepaspectratio]{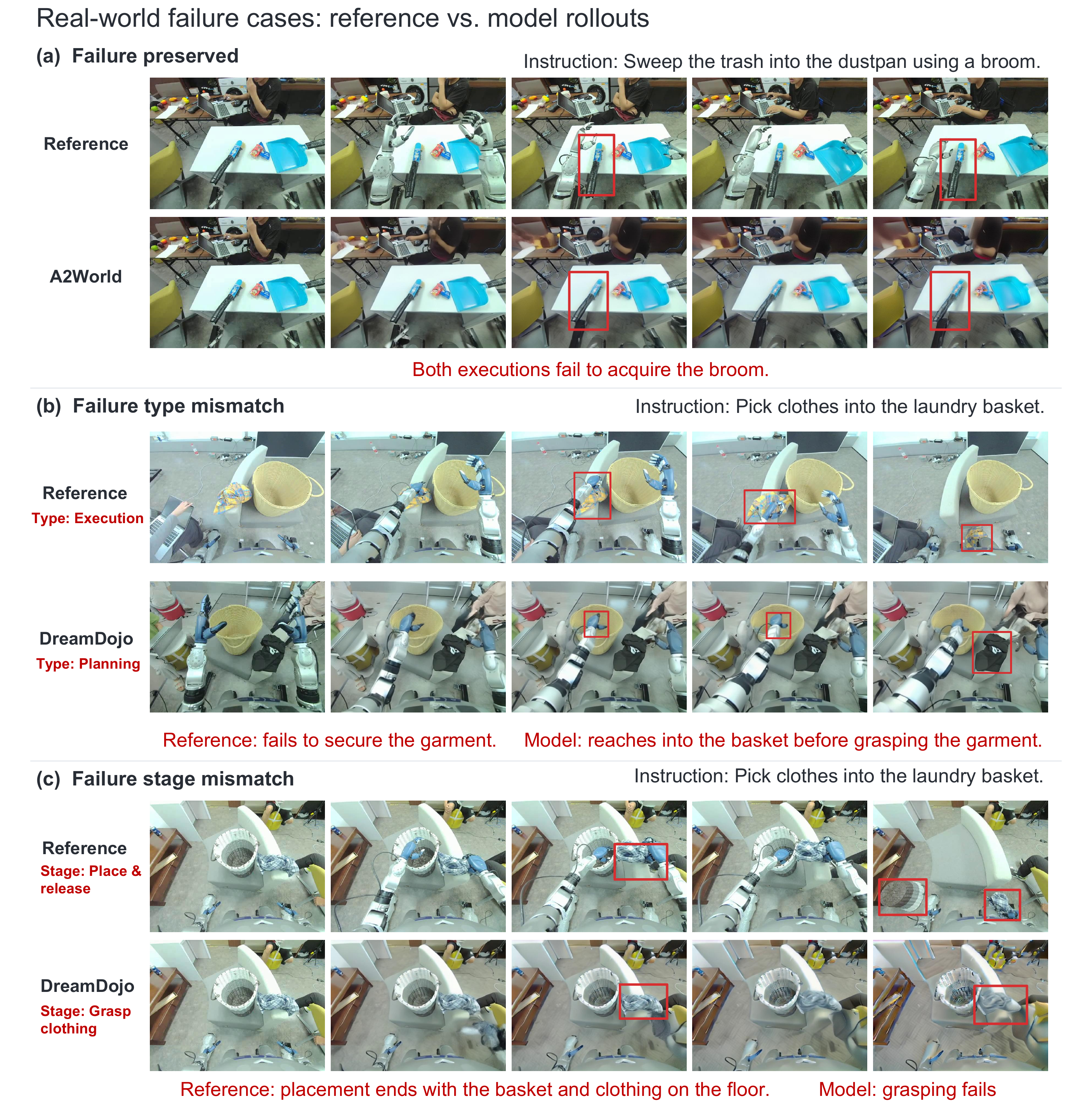}
\caption{Reference and world-model rollouts for real-world tasks on G1, illustrating (a) preserved failure, (b) a failure-type mismatch, and (c) a failure-stage mismatch. Frames proceed from left to right; instructions, labels, and visual evidence appear in each panel.}
\label{fig:g1_failure_comparisons}
\end{figure}

Figure~\ref{fig:g1_failure_comparisons} compares reference executions with A2World and DreamDojo on sweeping trash and collecting clothes.

\FloatBarrier

\noindent\begin{minipage}{\linewidth}
\section*{Reproducibility Statement}
\begingroup
\raggedright
Task and experimental settings, annotation, and boundary construction are specified in \mbox{Appendices~\ref{app:experimental_details}--\ref{app:boundary}}. Section~\ref{sec:protocol} defines the metrics.
\par
\endgroup

\vspace{6pt}
\section*{AI Use Statement}
AI assistance was used to review related work, reconcile experimental outputs, and prepare the manuscript and figures. Doubao-Seed-2-1-Pro and ChatGPT 6 were used for rollout outcome annotation, with expert adjudication of disagreements and random verification of agreed labels.
\end{minipage}\par

\end{document}